\documentclass{isprs}
\usepackage{subfigure}
\usepackage{setspace}
\usepackage{geometry}
\usepackage{epstopdf}
\usepackage[labelsep=period]{caption}
\usepackage[british]{babel} 
\usepackage[hang]{footmisc}
\usepackage{enumitem}

\usepackage{fix-cm}
\usepackage[T1]{fontenc}
\usepackage[hyphens]{url}

\usepackage[authoryear]{natbib}

\usepackage{lipsum}
\usepackage{amsmath}
\usepackage[colorlinks=true, linkcolor=blue, citecolor=blue, urlcolor=blue]{hyperref}
\usepackage[capitalize]{cleveref}
\usepackage[table]{xcolor}
\usepackage{amssymb}  
\usepackage{textcomp} 
\usepackage[title]{appendix}%
\usepackage{multirow}
\usepackage{booktabs}
\usepackage{makecell}
\usepackage{nicefrac}

\setcellgapes{5pt} 
\AtBeginEnvironment{appendices}{\crefalias{section}{appendix}} 

\newcommand{\squarecolor}[1]{\textcolor[HTML]{#1}{\rule{1.5ex}{1.5ex}}} 

\definecolor{firstplace}{RGB}{193,226,202}
\definecolor{secondplace}{RGB}{226,237,185}  
\definecolor{thirdplace}{RGB}{255, 250, 193}   

\begin{document}

\pagestyle{plain} 

\title{The Potential of Copernicus Satellites for Disaster Response:\\ Retrieving Building Damage from Sentinel-1 and Sentinel-2}
\date{}

\author{
    \parbox{\linewidth}{\centering
        Olivier Dietrich\textsuperscript{1},
        Merlin Alfredsson\textsuperscript{1},
        Emilia Arens\textsuperscript{2},
        Nando Metzger\textsuperscript{1},\\
        Torben Peters\textsuperscript{1},
        Linus Scheibenreif\textsuperscript{1},
        Jan Dirk Wegner\textsuperscript{2},
        Konrad Schindler\textsuperscript{1}%
    }%
}

\address{
	\textsuperscript{1 }Photogrammetry \& Remote Sensing Lab, ETH Zurich\\
	\textsuperscript{2 }EcoVision Lab, Department of Mathematical Modeling and Machine Learning (DM3L), University of Zurich
}

\abstract{
    Natural disasters demand rapid damage assessment to guide humanitarian response. Here, we investigate whether medium-resolution Earth observation images from the Copernicus program can support building damage assessment, complementing very-high resolution imagery with often limited availability. We introduce xBD-S12, a dataset of 10,315 pre- and post-disaster image pairs from both Sentinel-1 and Sentinel-2, spatially and temporally aligned with the established xBD benchmark. In a series of experiments, we demonstrate that building damage can be detected and mapped rather well in many disaster scenarios, despite the moderate 10$\,$m ground sampling distance. We also find that, for damage mapping at that resolution, architectural sophistication does not seem to bring much advantage: more complex model architectures tend to struggle with generalization to unseen disasters, and geospatial foundation models bring little practical benefit. Our results suggest that Copernicus images are a viable data source for rapid, wide-area damage assessment and could play an important role alongside VHR imagery. We release the xBD-S12 dataset, code, and trained models to support further research at \url{https://github.com/prs-eth/xbd-s12}.
}


\maketitle

\section{Introduction}\label{sec:intro}

Large-scale natural disasters like earthquakes, floods, and wildfires pose a persistent threat to large parts of humanity. In 2025 alone, events like the monsoon floods in Pakistan~\citep{mishra2025}, wildfires in California~\citep{mckoy2025}, and hurricanes in the Caribbean~\citep{melissa2025} have devastated entire communities in hours and have caused hundreds of casualties. Moreover, several types of disasters are induced by weather extremes and are intensifying as a consequence of climate change~\citep{van2006impacts, banholzer2014impact, IPCC_2023}.
Independent of their cause, all such events mandate an immediate humanitarian response. Yet, effective support requires information about the location and extent of damages. In recent years, satellite imagery has emerged as an important information source for disaster management.  Beyond rapidly providing an overview of the affected region, very-high resolution (VHR) images with ground sampling distances $\leq$2$\,$m have been shown to enable the identification of damaged assets, especially buildings, complementing sources on the ground~\citep{kawasaki2013growing, Rolla2025}.

Manually identifying damaged buildings in VHR imagery remains prohibitively time-consuming for fast disaster response. The machine learning revolution has naturally motivated efforts to automate this task~\citep{sun2020applications, braik2024automated}. Early approaches focused on single-hazard scenarios, e.g., tsunamis~\citep{fujita2017damage}, floods~\citep{rudner2019multi3net}, earthquakes~\citep{xu2019building}, hurricanes~\citep{cao2020building}, or wildfires~\citep{galanis2021damagemap}, successfully showing that deep neural networks can detect hazard-specific damage patterns in satellite imagery. Important progress came with the xView2 challenge and the accompanying xBD dataset~\citep{gupta2019creating, gupta2019xbd}, which enabled multi-hazard damage assessment for the first time. This benchmark catalyzed extensive research into automated damage assessment methodologies. The competition's winning solution employed a Siamese U-Net architecture with multiple encoder backbones, decomposing the problem into sequential building localization and damage classification stages~\citep{Durnov2019xview2}. This two-stage paradigm was subsequently adopted by several methods~\citep{zhao2020building, wu2021building}, while alternative approaches explored end-to-end joint optimization of both tasks through multi-task learning~\citep{weber2020building, gupta2021rescuenet, hao2021attention, zheng2021building}. More recently, the field has witnessed the integration of advanced architectures originally developed for other computer vision applications, including high-res networks~\citep[HRNet;][]{liu2022novel}, transformers~\citep{chen2022dual, kaur2023large}, and deep state-space models~\citep{chen2024changemamba}.

\begin{figure}[t!]
    \centering
    \includegraphics[width=.98\columnwidth]{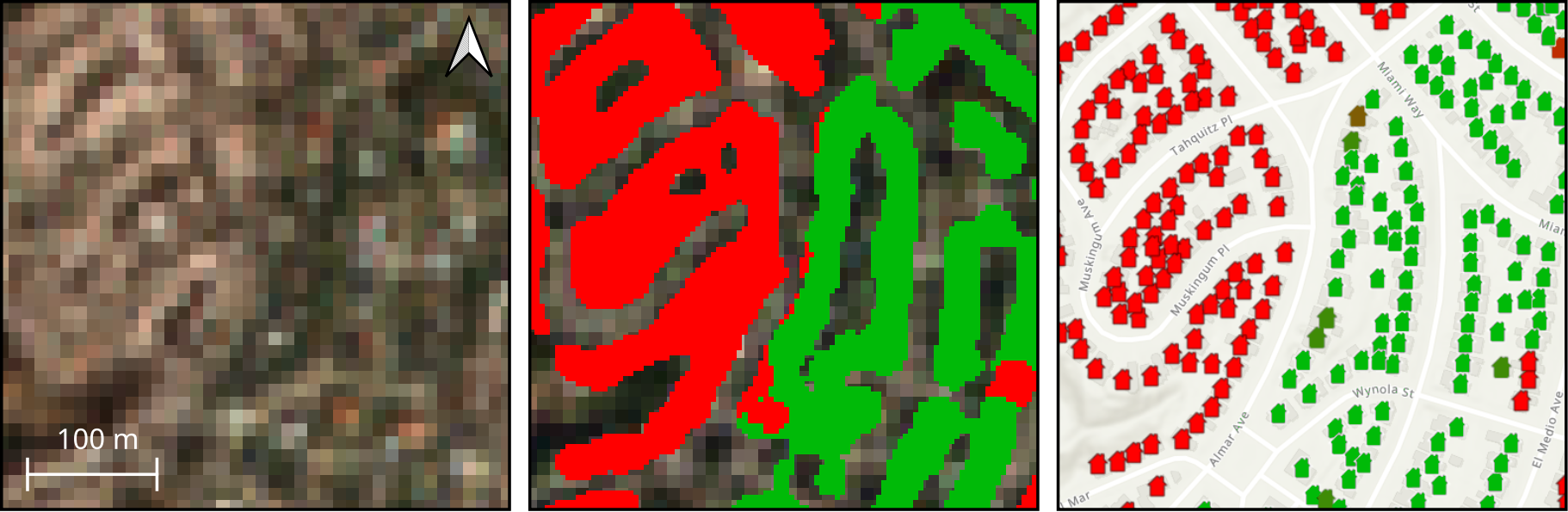}
    \vspace{-0.5em}
    \caption{Copernicus satellites are surprisingly effective at mapping building damage, despite the moderate GSD of 10$\,$m.  \underline{Left:} Sentinel-2 TCI product. \underline{Center:} zero-shot prediction from Sentinel-1 and Sentinel-2. \underline{Right:} reference map, adapted from \cite{palisadeFireMaps}.}
    \label{fig:teaser}
\end{figure}

Beyond architectural innovations, the research community has expanded the scope of available benchmarks to improve the diversity of imaging modalities and address operational constraints. Notable examples include the BRIGHT dataset~\citep{chen2025bright}, which combines pre-event VHR optical imagery with post-event VHR synthetic aperture radar (SAR) imagery, and the following DisasterM3, which combines xBD, BRIGHT, and ten new disaster events with textual descriptions for visual reasoning tasks~\citep{wang2025disasterm3}. In parallel, the broader remote sensing community has embraced geospatial foundation models (GeoFMs). Large, pre-trained models like Clay~\citep{Clay2023}, DOFA \citep{xiong2024neural}, Prithvi~\citep{szwarcman2024prithvi}, or AnySAT~\citep{astruc2025anysat} promise generalizable representations, learned from vast amounts of unlabeled satellite data. However, recent empirical studies suggest that they do not consistently attain the performance of task-specific, fully supervised baselines~\citep[e.g.,][]{corley2024change}. Vision-language models (VLMs) are another emerging frontier that could change how we interact with Earth observation data. Early systems such as TEOChat~\citep{irvin2024teochat} demonstrate the potential of natural language interfaces to interpret satellite images, yet remote sensing VLMs remain in their infancy.

While VHR imagery provides valuable detail for damage assessment and is often made public in the case of a natural disaster, its accessibility might be limited, which can be an issue especially during larger disasters~\citep{ainscoe2025earthquake}. Here, we investigate whether free, globally accessible satellite imagery can complement VHR data to perform building damage assessment at scale. Specifically, we explore the potential of the Copernicus missions, namely the multispectral imagery from Sentinel-2 and synthetic aperture radar (SAR) data from Sentinel-1, which both offer global coverage with sub-weekly revisits. To evaluate them for building damage assessment, we extend the established xBD dataset by pairing each VHR image with temporally aligned Sentinel-1/2 observations, enabling supervised learning of damage patterns at moderate resolution. Our contributions are: (1)~\textbf{xBD-S12}, a novel dataset that spatiotemporally aligns each xBD image pair with corresponding Sentinel-2 and Sentinel-1 acquisitions. As a byproduct, we provide corrected geolocations for the original xBD dataset. (2)~Deep learning models for damage assessment based on Sentinel imagery, accompanied by comprehensive per-disaster performance metrics to characterize model behavior across different event types. (3)~Ablation studies to examine the utility of GeoFMs for the task, demonstrating that they offer limited benefit. (4)~Independent, qualitative validation of the estimated damage maps for a disaster outside of xBD, by comparing to assessments from other sources. We release xBD-S12 as well as code and trained model weights, at \url{https://github.com/prs-eth/xbd-s12}.

\section{Material and Methods}\label{sec:methods}

\subsection{Data}
\begin{figure}[!thpb]
    \centering
    \includegraphics[width=.98\columnwidth]{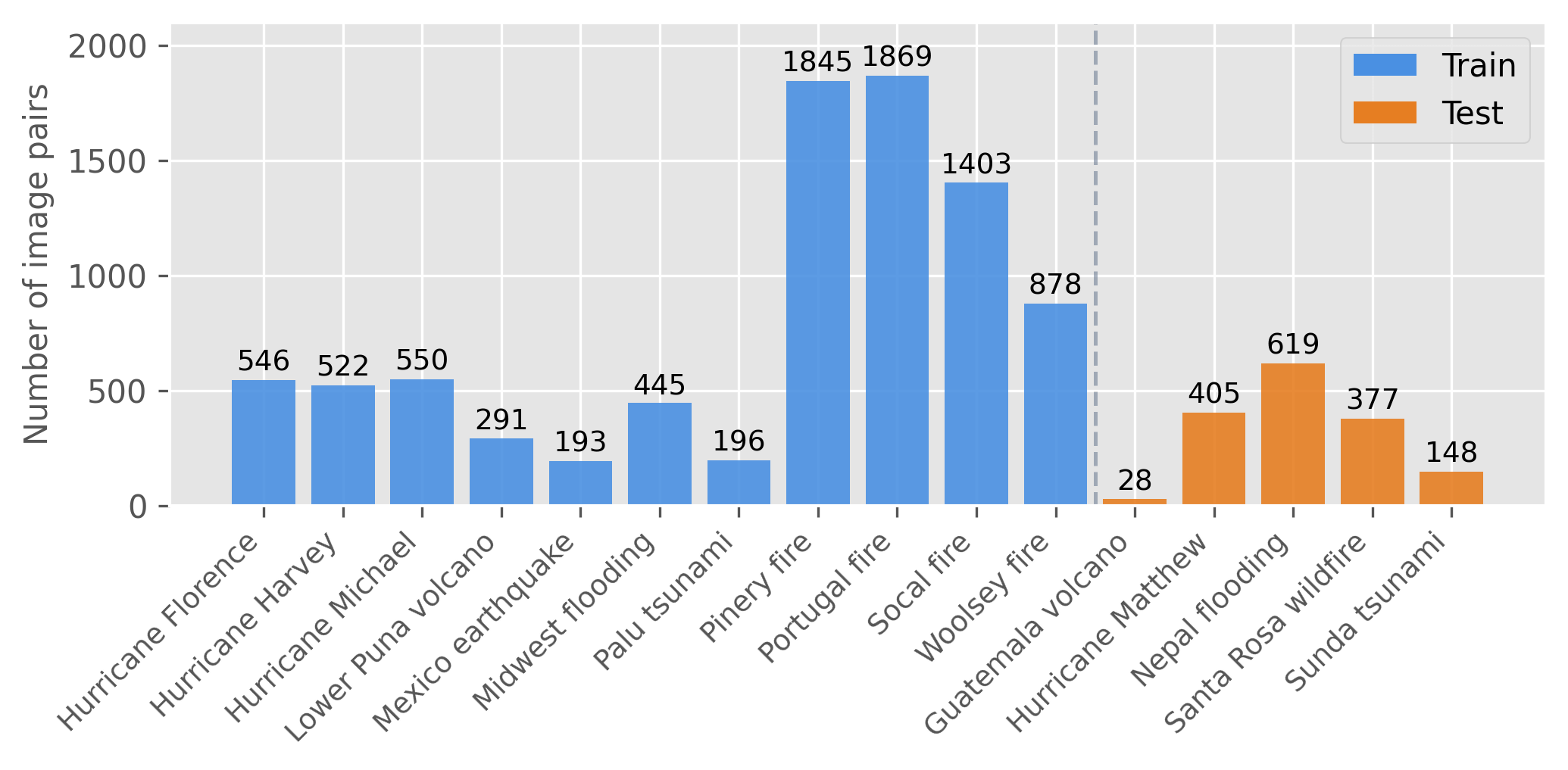}
    \vspace{-1em}
    \caption{xBD-S12 comprises 10,315 image pairs across 16 disaster events. Colours denote the \textit{event-based} split.}
    \label{fig:xbd_s12_distribution}
\end{figure}

\begin{figure}[!htpb]
    \centering
    \includegraphics[width=.95\columnwidth]{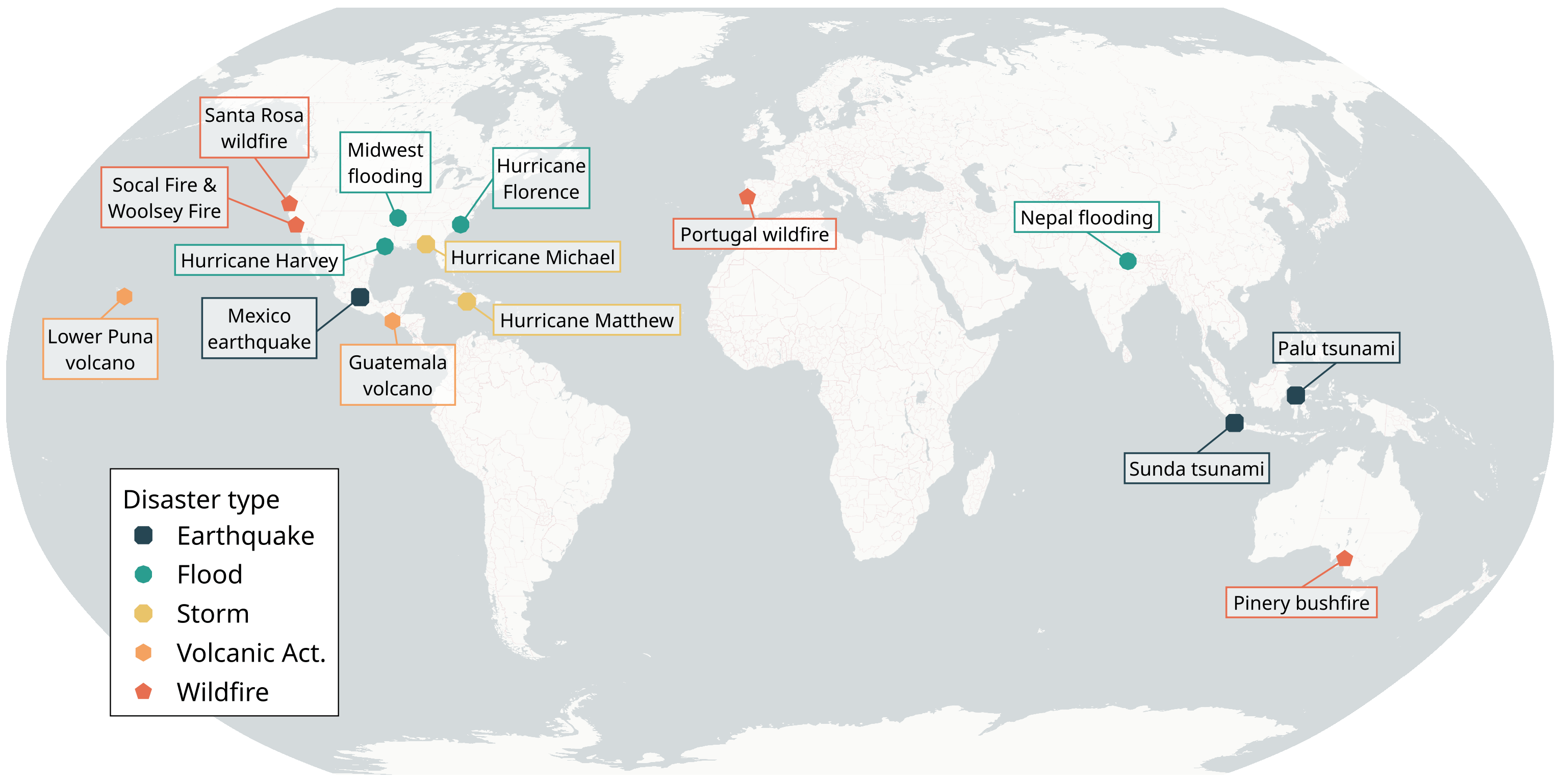}
    \vspace{-0.5em}
    \caption{Geographical distribution of xBD-S12, grouped by disaster type as introduced by \cite{DisasterAdaptiveNet25}.}
    \label{fig:xbd_overview}
\end{figure}

\begin{figure*}[!htpb]
    \centering
    \includegraphics[width=.9\textwidth]{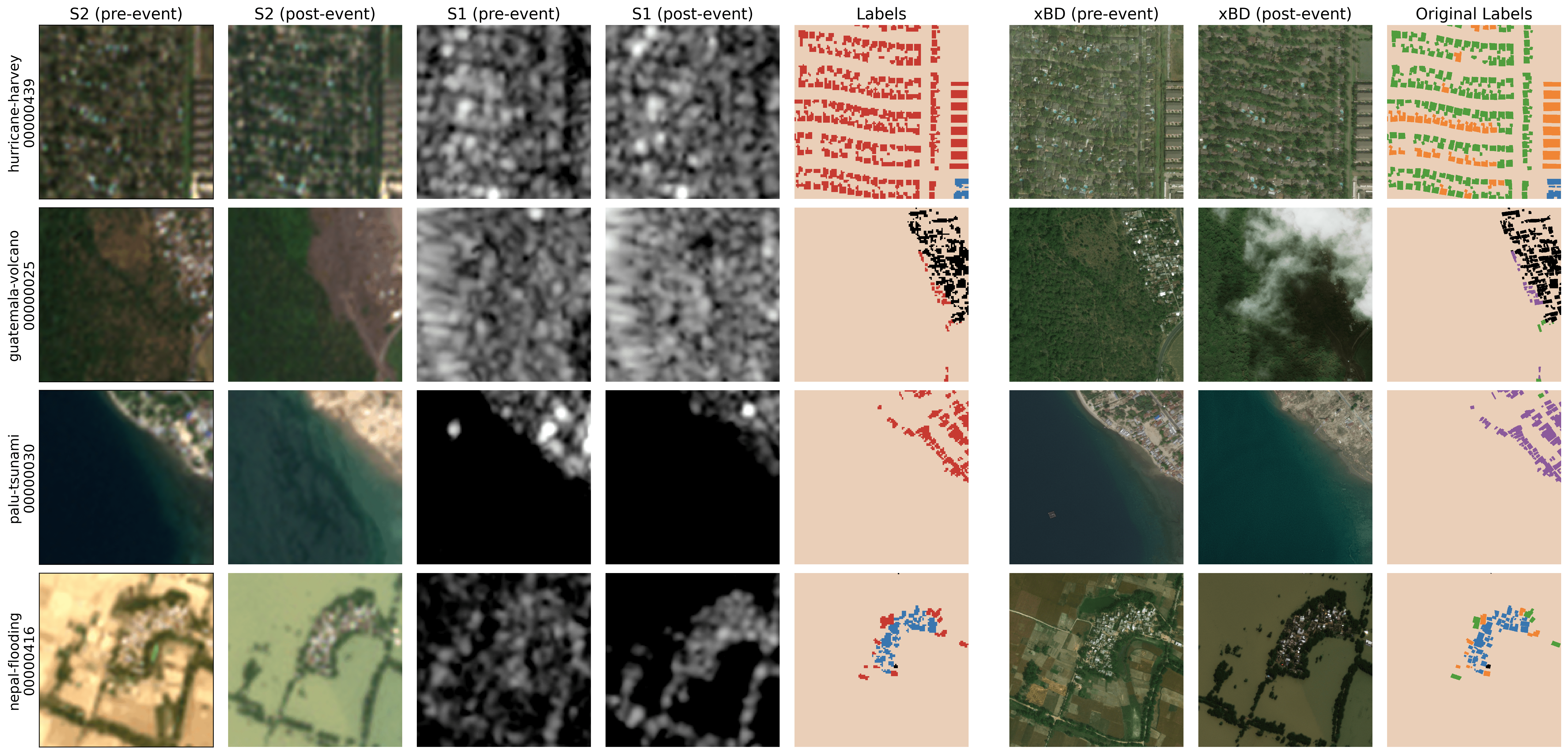}\\
    {\footnotesize
    \squarecolor{eacfb8} background\quad\squarecolor{000000} unknown\quad\squarecolor{3976af} undamaged\quad\squarecolor{c73a31} damaged\quad(\quad\squarecolor{f08535} \textcolor{gray}{minor damage}\quad\squarecolor{509d3d} \textcolor{gray}{major damage}\quad\squarecolor{8B5A9B} \textcolor{gray}{destroyed}\quad)
    }
    \caption{Example patches from xBD-S12. For visualisation purposes, we display the True Color Image product for Sentinel-2 and the VV-polarised (log-)amplitude for Sentinel-1. All tiles are 128$\times$128$\,$px ($\approx$4$\,$m GSD). On the right, VHR images (1024$\times$1024$\,$px, $\approx$0.5$\,$m GSD) and labels from the original high-resolution xBD dataset are shown for reference. See \cref{app:additional_visu} for more examples.}
    \label{fig:xbd_s12_examples}
\end{figure*}

We start by describing the construction of the xBD-S12 dataset. The data is summarized in \cref{fig:xbd_s12_distribution}, the geographical distribution is depicted in \cref{fig:xbd_overview}. and example patches are shown in \cref{fig:xbd_s12_examples}.

\subsubsection{Original xBD dataset.}
The basis of our work is the xBD dataset~\citep{gupta2019creating, gupta2019xbd}, originally developed for the xView2 challenge. It consists of 11,034 pre-/post-desaster VHR image pairs (each 1024×1024 pixels at $\approx$50$\,$cm/pixel) and over 425,000 building polygons. The image pairs are derived from 66 tiles released publicly by MAXAR (either through their Open Data Program or specifically for the challenge) and span 19 disaster events across seven countries. Events are grouped into five categories: earthquakes, floods, storms, volcanic activity, and wildfires. From the original tiles, the authors selected the relevant patches and contracted a commercial service to manually annotate building footprints and corresponding damage levels. Each building footprint is assigned one of four discrete damage classes: no damage, minor damage, major damage, or destroyed. Owing to its scale and high-quality annotations, the xBD dataset has become a well-established benchmark for post-disaster building damage assessment and remains widely used in the field.

The original dataset is divided into four subsets: \emph{train}, \emph{tier3}, \emph{test}, and \emph{holdout}. The train, test, and holdout sets represent 80/10/10\% random splits from the same ten disaster events, while the tier3 set contains data from nine additional disasters. During the challenge, participants trained their models on the train and tier3 sets, and used the test and holdout sets for evaluation. Due to the random split, the models see examples from every individual disaster location during training, with the associated danger of learning location- or event-specific patterns that do not generalize. To circumvent this,~\cite{DisasterAdaptiveNet25} proposed a different, event-based split that ensures geographical separation between the training and test sets, while maintaining at least one instance of each disaster type in the test set.

We systematically train and test our models with both splits to measure expected performance under two plausible scenarios: (1)~\textit{known distribution}, where a model is trained on a small labeled subregion of the specific event and then used to scale to the full region of interest, and (2)~\textit{unknown distribution}, where the model is applied to an unseen disaster for which no labels exist. Henceforth, we refer to the original split as \textit{xView2}, using the combined train and tier3 sets for training and the test set for evaluation. The alternative split of~\cite{DisasterAdaptiveNet25} is termed as \textit{event-based}.

Finally, the images in the xBD dataset suffer from georeferencing errors, likely due to an erroneous transformation. The misalignment does not matter for the original VHR task, which operates in image coordinates. It does, however, matter when aligning with Sentinel images for xBD-S12. We correct the coregistration, using the building footprints as ground control to fit an affine transformation. For details, see~\cref{app:align_xbd}.

\subsubsection{Sentinel images.}

Next, we determine the spatial extent of the coregistered xBD image patches, identify the corresponding Sentinel-1 and Sentinel-2 tiles, and download them, applying a consistent patch-level logic across all image pairs. Sentinel-2 tiles are selected based on cloud coverage and on temporal proximity to both the VHR images and the dates of the disasters. We rely on Google’s CloudScore+~\citep{CloudScorePlus} for cloud coverage, which has been precomputed for the entire Sentinel-2 archive and is accessible through Google Earth Engine~\citep[GEE,][]{Gorelick2017}. We make sure that post-event imagery clearly captures the disaster effects (e.g., flooding is only visible on a few days after the event), whereas we allow for a broader date range in the pre-event imagery and optimize for minimal cloud cover. To facilitate access to True Color Image (TCI) products, we download the original \textit{.SAFE} Sentinel-2 Level-2A files using \texttt{$\Phi$}-down~\citep{delprete2025phidown}. The level-2A product consists of 12 multispectral bands with native ground sampling distances (GSDs) ranging from 10$\,$m to 60$\,$m. 

Once Sentinel-2 images are selected, we identify the closest Sentinel-1 acquisitions, again with respect to both the VHR images and the disaster dates. We also require the pre- and post-event Sentinel-1 images to share the same orbit direction, since ascending and descending orbits lead to very different viewing geometries of the same area. We do not prioritize one orbit direction over the other, resulting in a mix of both across the dataset. For one disaster (Hurricane Matthew), pre- and post-disaster images from the same orbit direction are not available, so we allow images with different directions. We use the Ground Range Detected (GRD) product available on GEE, which provides log-amplitudes in the VV and VH polarizations, resampled to 10$\,$m. GEE automatically performs standard preprocessing for each tile, namely thermal noise removal, radiometric calibration, and terrain correction. We do not perform any additional preprocessing. For more details on dataset creation, see~\cref{app:sentinel_details}.

\subsubsection{Preprocessing.}
We discard three disasters, namely the three tornado events, as they occurred before the launch of the Sentinel missions. For the remaining 16 disasters, we extract Sentinel-1 and Sentinel-2 patches that match the spatial extent of the xBD patches, resulting in a total of 10,315 image pairs.

\paragraph{Spatial resampling.} The xBD dataset features VHR images with slightly varying GSDs. We find that, translated to Sentinel's 10$\,$m resolution, this would result in patch sizes varying between 48 and 52 pixels. To obtain consistent input dimensions across our dataset, we resample all Sentinel patches to a fixed size of 128$\times$128 pixels with Lanczos interpolation, corresponding to an effective GSD of $\approx$4$\,$m. This resampling serves three purposes: (1)~it standardizes the patch dimensions in the xBD-S12 dataset, (2)~it provides a more suitable input size for modern deep learning architectures, and (3)~there is evidence that building detection in Sentinel data benefits from moderate upsampling~\citep{sirko2023high}. 

\paragraph{Label simplification.} The original fine-grained classification task is to some degree ill-posed, because the delimitation of the four damage classes is somewhat ambiguous and depends on the disaster type~\citep{gupta2019xbd, DisasterAdaptiveNet25}. What is more, at 10$\,$m GSD individual buildings occupy only few pixels, so that the distinction becomes impractical. We therefore merge all damage classes (minor, major, destroyed) into a single \emph{damaged} class, resulting in three final classes: \emph{background}, \emph{intact}, and \emph{damaged}. This simplification aligns better with what can realistically be expected from damage mapping with Sentinel data, where the aim is to identify affected buildings rather than to quantify the exact degree of damage.

\paragraph{Invalid mask.} We create an \textit{invalid} mask by keeping track of unclassified pixels in the original xBD dataset. These pixels occur either when the VHR image does not cover the full patch or when the post-event image is cloudy. These pixels are excluded from training and evaluation.

\subsection{Models}

\paragraph{Problem formulation.}
\label{sec:problem}
We formulate building damage assessment as a multiclass semantic segmentation problem. Our input consists of pre- and post-disaster images from Sentinel-1 and Sentinel-2: $\mathbf{X}_{\text{S1,pre}}$, $\mathbf{X}_{\text{S1,post}}$ $\in \mathbb{R}^{2 \times 128 \times 128}$ and $\mathbf{X}_{\text{S2,pre}}$, $\mathbf{X}_{\text{S2,post}}$ $\in \mathbb{R}^{12 \times 128 \times 128}$. The task is to predict a damage map $\hat{\mathbf{Y}}$ $\in \mathbb{R}^{128 \times 128}$ where each pixel belongs to one of the three classes:  \textit{background}, \textit{intact}, or \textit{damaged}.

\subsubsection{Architecture.}
Our baseline is a standard U-Net with ResNet34 backbone, pretrained on ImageNet, followed by a lightweight segmentation head that predicts per-pixel class logits. Through ablation studies, we found that an encoder depth of three layers is sufficient and that removing all batch normalization layers from the encoder improves performance.

\paragraph{Early vs.\ late fusion.} We test two fusion strategies. In the \textit{early fusion}, all pre- and post-event bands are concatenated channel-wise and fed directly to a single U-Net. In the \textit{late fusion}, we use a Siamese design to separately encode the pre- and post-event images through a shared U-Net (without the segmentation head). The resulting feature maps are then concatenated before the final segmentation head.

\paragraph{Two-step vs.\ joint optimization.} We test two variants: either a single model that directly predicts the three classes (\textit{joint}); or two separate models (\textit{2-step}), one that localizes buildings via binary segmentation into building vs.\ background, and a second one that performs binary damage classification. In the latter case, the final damage map is obtained by masking the damage predictions to the estimated building pixels.

\paragraph{Ensembling.} For robustness, we perform model ensembling by averaging the logits of three networks trained with different seeds before converting them to class probabilities with the final activation function.

\subsubsection{Comparison methods.}
We compare our straightforward segmentation network against multiple dedicated architectures developed for damage mapping in VHR imagery:

\begin{itemize}[label=-,leftmargin=*,itemsep=0pt]
\item
\textit{Strong baseline}~\citep{DisasterAdaptiveNet25}: A simplified version of the xView2 winning solution, framed as multiple independent binary semantic segmentations.
\item
\textit{DisasterAdaptiveNet}~\citep{DisasterAdaptiveNet25}: An extension of the strong baseline that adds a FiLM~\citep{perez2018film} module to modulate the predictions according to the disaster type.
\item
\textit{ChangeOS}~\citep{zheng2021building}:
A deep object-based semantic change detection framework using a partial Siamese encoder with task-aware encoders and decoders.
\item
\textit{ChangeMamba}~\citep{chen2024changemamba}:
Change detection based on the Mamba~\citep{gu2024mamba} deep state space architecture, designed to enable the global receptive field of Transformers with a lower memory footprint.
\end{itemize}

\subsubsection{Foundation models.}
We also test two recent geospatial \textit{foundation} models: Prithvi-EO-2.0 with 300M parameters~\citep{szwarcman2024prithvi}, as available in Terratorch~\citep{gomes2025terratorch}, and DOFA~\citep{xiong2024neural} with ViT-Base backbone, as implemented in TorchGeo~\citep{Stewart_TorchGeo_Deep_Learning_2024}. In both cases, we keep the backbone frozen and only finetune a UperNet~\citep{xiao2018unified} to decode the features into segmentation masks. We experiment with training either on the full dataset or on a small subset to simulate a low-data regime.

\subsection{Training and evaluation}

\subsubsection{Building footprint buffering.}
\label{sec:buffer}
We apply a morphological dilation to obtain a 3-pixel buffer around the ground truth building footprints and add the buffer pixels to the \textit{invalid} mask during training. This strategy (1) mitigates potential misregistration between MAXAR and Sentinel imagery, and (2) encourages bolder predictions by not penalizing outputs extending slightly beyond ground truth footprints. Given Sentinel's medium resolution, precise boundary delineation is challenging; we therefore prioritize recall over boundary alignment and accepting slightly inflated predictions. During inference, we evaluate both with and without the buffer. A performance drop when removing the buffer from the evaluation indicates that the model indeed inflated the footprints, in practice an acceptable price to pay for more complete damage detection.

\subsubsection{Performance metrics.}
Following the xView2 Challenge, we separately evaluate the results w.r.t.\ building localization and damage detection, in both cases using the F1 score (the harmonic mean between precision and recall).
We report two scores: the localization score F1\textsubscript{loc} and the damage score F1\textsubscript{dmg}. Unless stated otherwise, F1\textsubscript{loc} is evaluated without a buffer around the building footprints. We additionally show F1\textsubscript{loc, B=X}, evaluated with an X-pixel buffer around the footprints. The damage score F1\textsubscript{dmg} is the harmonic mean between the F1 scores of classes \textit{intact} and \textit{damaged}. Following the xView2 challenge, it is computed only over pixels within the ground truth building footprints, so F1\textsubscript{dmg} is not impacted by the buffer.
Again following xView2, the localization and damage scores are linearly combined into an overall ``competition score'':  $\text{F1}_\text{comp} = 0.3\cdot \text{F1}_\text{loc} + 0.7\cdot \text{F1}_\text{dmg}$.

\subsubsection{Training details.}
\label{sec:training_details}
Sentinel-1 and Sentinel-2 data are normalized channel-wise using percentile-based min-max scaling: $x_{\text{norm}} = \text{clip}(\nicefrac{(x - p_1)}{(p_{99} - p_1)}, 0, 1)$, where $p_1$ and $p_{99}$ are computed from the training set. During training, we apply geometric augmentations (random horizontal/vertical flips and 90-degree rotations) jointly to imagery and labels.

To address class imbalance (background pixels vastly outnumber building pixels, and intact buildings outnumber damaged ones), we employ biased random sampling, where the probability of drawing a sample is inversely proportional to the frequency of its pixel classes in the training set.

All models are trained for 40 epochs using the AdamW optimizer~\citep{loshchilov2017decoupled} with learning rate $5 \times 10^{-4}$, weight decay $1 \times 10^{-4}$, and cosine annealing schedule with 3 epochs of linear warm-up. Unlike prior works that employ specialized loss functions, we found the standard cross-entropy loss to perform best, likely due to our simplified label set. As the final model to be evaluated on the test set, we retain the checkpoint with highest F1\textsubscript{comp} on the validation set (respectively higher F1\textsubscript{loc} for localization-only experiments).

\section{Results}\label{sec:results}

\begin{table*}[!htpb]
\centering
\footnotesize
\begin{tabular}{@{}lcccccc@{}}
\toprule
& \multicolumn{3}{c}{\textit{xView2 split}} & \multicolumn{3}{c}{\textit{event-based split}}\\
\cmidrule(lr){2-4} \cmidrule(lr){5-7}
Model & F1\textsubscript{comp (B=3)} & F1\textsubscript{loc (B=3)} & F1\textsubscript{dmg} & F1\textsubscript{comp (B=3)} & F1\textsubscript{loc (B=3)} & F1\textsubscript{dmg} \\
\midrule
U-Net (2-step, early fusion) & 0.760 {\scriptsize (0.848)} & 0.597 {\scriptsize (0.891)} & \cellcolor{thirdplace}{0.830} & \cellcolor{thirdplace}{0.690 {\scriptsize (0.756)}} & \cellcolor{firstplace}{\textbf{0.502} {\scriptsize (0.722)}} & 0.771\\ \addlinespace[1pt]
U-Net (2-step, late fusion) & \cellcolor{thirdplace}{0.761 {\scriptsize (0.849)}} & 0.597 {\scriptsize (0.891)} & \cellcolor{secondplace}{0.831} & \cellcolor{firstplace}{\textbf{0.710} {\scriptsize (0.775)}} & \cellcolor{firstplace}{\textbf{0.502} {\scriptsize (0.722)}} & \cellcolor{firstplace}{\textbf{0.799}}\\ \addlinespace[1pt]
U-Net (joint, early fusion) & \cellcolor{secondplace}{0.764 {\scriptsize (0.843)}} & \cellcolor{secondplace}{0.611 {\scriptsize (0.873)}} & \cellcolor{thirdplace}{0.830} & 0.687 {\scriptsize (0.738)} & \cellcolor{secondplace}{0.491 {\scriptsize (0.662)}} & 0.771\\ \addlinespace[1pt]
U-Net (joint, late fusion) & 0.760 {\scriptsize (0.845)} & \cellcolor{thirdplace}{0.606 {\scriptsize (0.888)}} & 0.826 & \cellcolor{secondplace}{0.692 {\scriptsize (0.743)}} & \cellcolor{thirdplace}{0.455 {\scriptsize (0.622)}} & \cellcolor{thirdplace}{0.794}\\ \addlinespace[1pt]
StrongBaseline & 0.760 {\scriptsize (0.844)} & 0.595 {\scriptsize (0.876)} & \cellcolor{secondplace}{0.831} & \cellcolor{thirdplace}{0.690 {\scriptsize (0.740)}} & 0.444 {\scriptsize (0.609)} & \cellcolor{secondplace}{0.796}\\ \addlinespace[1pt]
DisasterAdaptiveNet & 0.734 {\scriptsize (0.815)} & 0.599 {\scriptsize (0.869)} & 0.792 & 0.636 {\scriptsize (0.655)} & 0.358 {\scriptsize (0.419)} & 0.756\\ \addlinespace[1pt]
ChangeOS & 0.718 {\scriptsize (0.801)} & 0.488 {\scriptsize (0.767)} & 0.816 & 0.589 {\scriptsize (0.644)} & 0.432 {\scriptsize (0.615)} & 0.656\\ \addlinespace[1pt]
ChangeMamba & \cellcolor{firstplace}{\textbf{0.800} {\scriptsize (0.831)}} & \cellcolor{firstplace}{\textbf{0.651} {\scriptsize (0.755)}} & \cellcolor{firstplace}{\textbf{0.863}} & 0.655 {\scriptsize (0.667)} & 0.362 {\scriptsize (0.402)} & 0.780\\
\bottomrule
\end{tabular}
\vspace{-0.5em}
\caption{F1 scores for different approaches on both the original xView2 split and the event-based split proposed by~\cite{DisasterAdaptiveNet25}. Values are computed on an ensemble of three runs where logits were averaged before thresholding. Values in parentheses are computed excluding a 3-pixel buffer around buildings during evaluation. Since F1\textsubscript{dmg} is computed only on the building pixels in the groundtruth, it is not affected by the buffer. We highlight the top three results for each metric as \colorbox{firstplace}{\textbf{first}}, \colorbox{secondplace}{second}, and \colorbox{thirdplace}{third}.}
\label{tab:main_results}
\end{table*}

\cref{tab:main_results} presents the detailed results of our approach against several SOTA models. We report metrics both with and without a 3-pixel buffer around the building footprints. From these metrics, we make several key observations:

\begin{itemize}[label=-,leftmargin=*,itemsep=0pt]
\item ChangeMamba achieves the best performance on the original xView2 split, but exhibits poor generalization to unseen disasters in the event-based split. This suggests that the model memorizes event- or location-specific characteristics from the training data, which then hamper generalization to unseen events and locations at inference.
    
\item In contrast, our simpler two-step approach works best for the event-based split. We hypothesize that, with its lower capacity, it operates more like a conventional change detection model and captures fewer intricate, event-specific patterns; thus generalizing better to unseen events. The result is in line with the general trend that architectural sophistication becomes less important as the GSD increases. 

\item Joint optimization is beneficial when training and testing in the same distribution (xView2 split), but the less elegant 2-step approach is more robust when mapping unseen disasters (event-based split).

\item The 3-pixel buffer greatly increases the localization score of all models, indicating that they detect buildings correctly and inflate their outline, as expected. In other words, training with the buffer has the desired effect of increasing per-building recall at the cost of per-pixel precision, so as to avoid missing possible damages.
    
\item The FiLM layer in DisasterAdaptiveNet provides no improvement over the baseline (StrongBaseline). It appears that conditioning on the disaster type either benefits only VHR image analysis or, more likely, that it improves the discrimination of detailed damage classes like \emph{minor} vs.\ \emph{major} damage, which we merged and whose definitions are disaster-dependent.
    
\item ChangeOS markedly underperforms compared to the other tested models. The reason is likely its object-based approach, which becomes less effective at the 10$\,$m resolution of Sentinel, where buildings and other man-made urban structures no longer form well-delineated, separable ``objects''.
\end{itemize}

For the remainder of this section, we continue with our best-performing model, i.e., the 2-step strategy with the Siamese (late fusion) architecture.

\subsection{Performance per disaster event}
\cref{tab:score_disaster} shows the metrics per event, for both test splits. There are substantial performance variations, highlighting both the capabilities and the limitations of medium-resolution satellite imagery for damage assessment.

In the in-domain xView2 split, we achieve relatively high scores on all disasters. The highest scores are reached for wildfires, floods, and tsunamis, which can be explained by the nature of the associated damage, where large contiguous areas are affected in a way that significantly alters their spectral signatures (burn scars, respectively water). A notable exception is the \emph{Mexico earthquake}, for which we cannot detect any damage. This disaster exposes a fundamental limitation of the lower spatial resolution: Damages are sparse and only affect individual, small buildings dispersed across a densely built-up area, rendering them invisible at the Sentinels' 10$\,$m GSD (also for human interpreters). See \cref{app:poor_performing_disasters} for more details on this disaster.

As expected, performance is in general lower for the event-based split, since the training did not include data from the vicinity of the test regions. Only for the \textit{Guatemala volcano}, performance increases, which may at first seem surprising. It turns out that the discrepancy stems from the particular choice of test data in the xView2 split. The test set for this small event consists of only five patches, all of which happen to depict challenging regions, with small buildings at the boundaries of lava streams and significant distribution shifts compared to the other patches from the same disaster.

\begin{table}[!htpb]
\centering
\footnotesize
\begin{tabular}{@{}l|lccc@{}}
\toprule
 \multicolumn{1}{c}{} 
& & F1\textsubscript{comp (B=3)} & F1\textsubscript{loc (B=3)} & F1\textsubscript{dmg} \\
\midrule
\multirow{10}{*}{\rotatebox{90}{\textit{xView2 split}}}
& Guatemala volcano & 0.611 {\scriptsize (0.671)} & 0.484 {\scriptsize (0.684)} & 0.665 \\
& Hurricane Florence & 0.785 {\scriptsize (0.852)} & 0.485 {\scriptsize (0.708)} & 0.913 \\
& Hurricane Harvey & 0.749 {\scriptsize (0.839)} & 0.616 {\scriptsize (0.917)} & 0.806 \\
& Hurricane Matthew & 0.557 {\scriptsize (0.628)} & 0.423 {\scriptsize (0.659)} & 0.615 \\
& Hurricane Michael & 0.596 {\scriptsize (0.683)} & 0.566 {\scriptsize (0.853)} & 0.609 \\
& Mexico earthquake & 0.194 {\scriptsize (0.287)} & 0.648 {\scriptsize (0.958)} & 0.000 \\
& Midwest flooding & 0.612 {\scriptsize (0.692)} & 0.530 {\scriptsize (0.795)} & 0.647 \\
& Palu tsunami & 0.777 {\scriptsize (0.863)} & 0.643 {\scriptsize (0.929)} & 0.834 \\
& Santa Rosa wildfire & 0.842 {\scriptsize (0.942)} & 0.575 {\scriptsize (0.909)} & 0.956 \\
& Socal fire & 0.763 {\scriptsize (0.843)} & 0.548 {\scriptsize (0.815)} & 0.855 \\
\midrule
\multirow{5}{*}{\rotatebox{90}{\textit{event-based}}}
& Guatemala volcano & 0.721 {\scriptsize (0.809)} & 0.467 {\scriptsize (0.760)} & 0.830 \\
& Hurricane Matthew & 0.403 {\scriptsize (0.474)} & 0.444 {\scriptsize (0.683)} & 0.385 \\
& Nepal flooding & 0.616 {\scriptsize (0.655)} & 0.441 {\scriptsize (0.572)} & 0.690 \\
& Santa Rosa wildfire & 0.747 {\scriptsize (0.828)} & 0.547 {\scriptsize (0.817)} & 0.833 \\
& Sunda tsunami & 0.336 {\scriptsize (0.396)} & 0.536 {\scriptsize (0.739)} & 0.249 \\
\bottomrule
\end{tabular}
\vspace{-0.5em}
\caption{Per-disaster performance breakdown for both test splits. Metrics are computed using the same methodology as \cref{tab:main_results}.}
\label{tab:score_disaster}
\end{table}

\subsection{Input data for building localization}
When using the 2-step approach, the pre-disaster image alone would, in principle, be sufficient for the building localization task. 
Nevertheless, feeding also post-event images---which must be available for damage mapping---should improve performance, if only by providing redundancy for undamaged (or lightly damaged) buildings. \cref{tab:loc_scores} shows F1\textsubscript{loc} scores for both the validation and test subsets.
They confirm that access to both pre- and post-event views consistently improves performance. The most significant gain occurs for the test set of the event-based split. We hypothesize that, by seeing views of the same buildings captured on different dates, the model learns some degree of invariance to atmospheric and lighting conditions, which is beneficial when applied to unseen geographic regions.

\begin{table}[!htpb]
\centering
\footnotesize
\begin{tabular}{@{}lcccc@{}}
\toprule
& \multicolumn{2}{c}{\textit{xView2 split}} & \multicolumn{2}{c}{\textit{event-based split}}\\ 
\cmidrule(lr){2-3} \cmidrule(lr){4-5}
Input & Valid & Test & Valid & Test \\ 
\midrule
Pre & 0.833 & 0.868 & 0.869 & 0.566\\\addlinespace[1pt]
Pre+Post & 0.865 & 0.891 & 0.891 & 0.722\\
\bottomrule
\end{tabular}
\vspace{-0.5em}
\caption{F1\textsubscript{loc (B=3)} for different input configurations. Multi-temporal inputs improve generalization to unseen locations, respectively events.}
\label{tab:loc_scores}
\end{table}

\subsection{Effect of buffer size}
\begin{figure}[!htpb]
    \centering
    \includegraphics[width=.90\columnwidth]{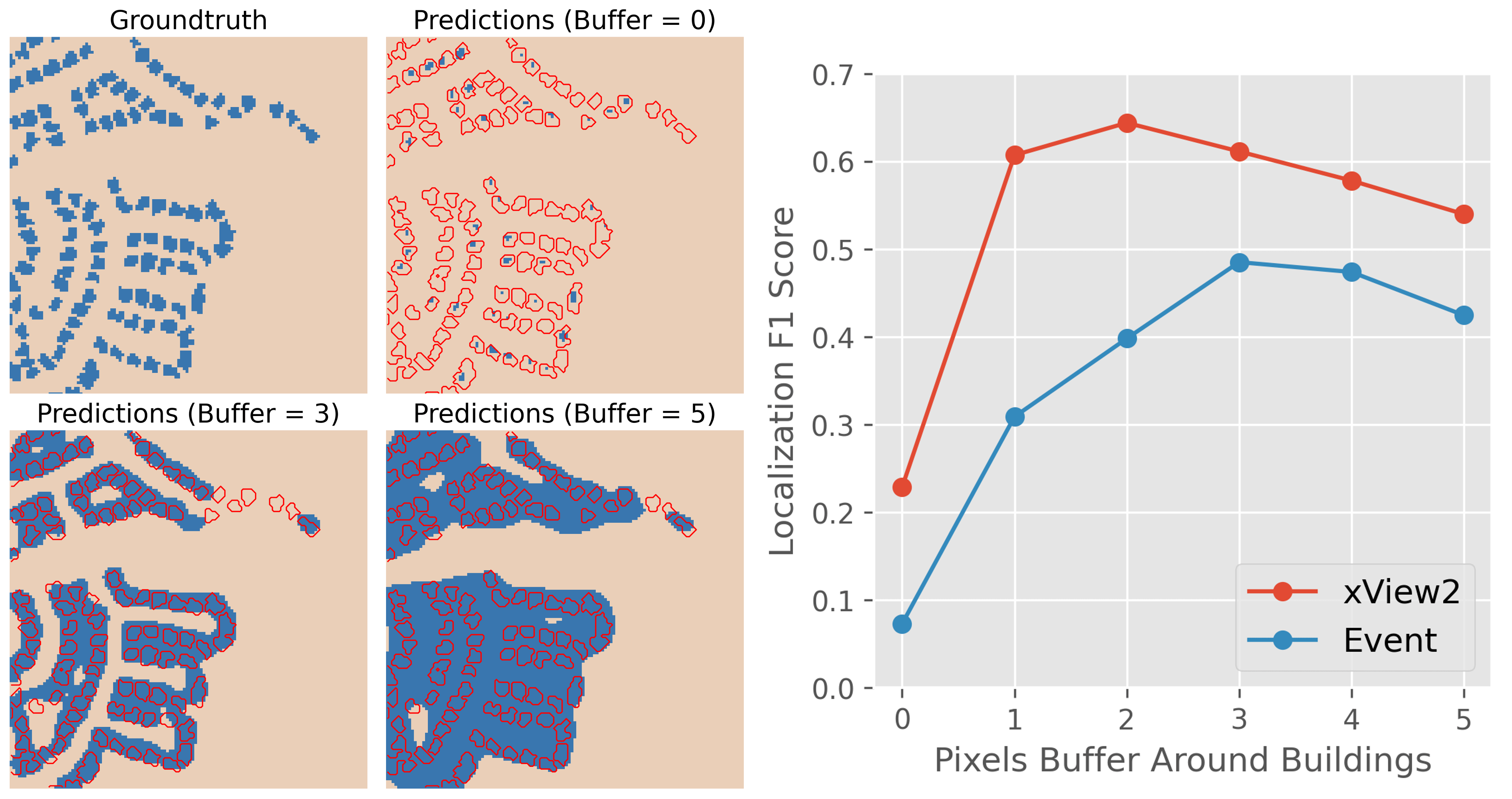}
    \vspace{-0.5em}
    \caption{Effect of buffer size on building localization.\\\underline{Left:} Predicted footprints with different training buffer sizes. \underline{Right:} F1\textsubscript{loc} score as a function of buffer size.}
    \label{fig:buffer_ablation}
\end{figure}

Masking a buffer around building footprints during training significantly improves mapping performance. To determine the right size for the buffer, we train models with buffers ranging from 0 to 5 pixels and compute the corresponding localization scores (\cref{fig:buffer_ablation}). As expected, larger buffers yield increasingly diffuse predictions, trading pixel-level accuracy for higher building-level recall. Depending on the data split, localization scores peak around 2 to 3 pixels. On average, the 3-pixel buffer enjoys a slight edge, hence we adopt it as our default setting. I.e., we accept inflated building outlines to minimize the risk of missing damaged buildings.

\subsection{Geospatial foundation models}
We additionally evaluate whether pretrained GeoFMs can improve damage assessment. We test two recent models, Prithvi-EO-2.0-300M~\citep{szwarcman2024prithvi} and DOFA-Base~\citep{xiong2024neural}. We treat the GeoFM as a frozen feature extractor and finetune only the UperNet decoder (see \cref{app:geofm_finetuning}). \cref{tab:geofms} summarizes results for four different scenarios.

In the in-domain xView2 split, Prithvi achieves a slightly higher overall score than our 2-step model (but still significantly lower than ChangeMamba, \cref{tab:main_results}), with DOFA not too far behind.
However, that performance plummets as one shifts to the event-based split. It appears that, when faced with a generic feature extractor not tuned to damage mapping, the decoder has a stronger tendency to overfit to the specific conditions of the training set, thus generalizing poorly to unseen events (respectively, locations).

To probe the value of GeoFMs in the low-data regime, we also train only on a single event, then test on another of the same type. The reasoning is that with so little training data, pretrained features should be particularly helpful.
Somewhat unexpectedly, for wildfires, this is not the case. All methods achieve high scores. In other words, burnt areas are well detectable at Sentinel resolution, but even a small amount of data is sufficient to learn that detection as well as a GeoFM. Overall, the U-Net maintains the best overall performance, driven by superior building localization.
For hurricanes, the two GeoFMs do achieve notable improvements over the U-Net baseline. Yet, absolute mapping performance remains very low for all models, indicating that more data is needed to train a meaningful building detector or damage classifier---even if the feature extractor has been pretrained on massive data. 

Overall, we conclude that pretrained GeoFMs are suitable for detecting broad environmental changes like burnt or flooded areas, but struggle with more fine-grained tasks like building localization.
Consequently, existing GeoFMs bring only small, if any practical benefits for damage mapping.
We point out that finetuning also the encoder (feature extractor) of a GeoFM might improve its performance. However, this would incur substantially higher computational cost than training the lightweight U-Net model from scratch.

\begin{table}[!htpb]
\centering
\footnotesize
\begin{tabular}{@{}l|lccc@{}}
\toprule
 & & F1\textsubscript{comp (B=3)} & F1\textsubscript{loc (B=3)} & F1\textsubscript{dmg} \\
\midrule
\textit{xView2 split} & U-Net & 0.761 {\scriptsize (0.849)} & \textbf{0.597} {\scriptsize (0.891)} & 0.831\\
& Prithvi & \textbf{0.764} {\scriptsize (0.840)} & 0.571 {\scriptsize (0.821)} & \textbf{0.847}\\
& DOFA & 0.723 {\scriptsize (0.797)} & 0.557 {\scriptsize (0.805)} & 0.794\\
\midrule
\textit{event-based split} & U-Net & \textbf{0.710} {\scriptsize (0.775)} & \textbf{0.502} {\scriptsize (0.722)} & \textbf{0.799}\\
& Prithvi & 0.634 {\scriptsize (0.695)} & 0.437 {\scriptsize (0.641)} & 0.718\\
& DOFA & 0.533 {\scriptsize (0.594)} & 0.436 {\scriptsize (0.642)} & 0.574\\
\midrule
\textit{{Woolsey fire \textrightarrow{}}} & U-Net & \textbf{0.746} {\scriptsize (0.784)} & \textbf{0.462} {\scriptsize (0.589)} & 0.868\\
\textit{Santa Rosa fire} & Prithvi & 0.725 {\scriptsize (0.745)} & 0.312 {\scriptsize (0.378)} & \textbf{0.902}\\
& DOFA & 0.698 {\scriptsize (0.760)} & 0.405 {\scriptsize (0.615)} & 0.823\\
\midrule
\textit{hurr.\ Michael \textrightarrow{}} & U-Net & 0.143 {\scriptsize (0.178)} & \textbf{0.349} {\scriptsize (0.464)} & 0.055\\
\textit{hurr.\ Matthew} & Prithvi & \textbf{0.271} {\scriptsize (0.316)} & 0.309 {\scriptsize (0.460)} & \textbf{0.255}\\
& DOFA & 0.255 {\scriptsize (0.293)} & 0.300 {\scriptsize (0.427)} & 0.236\\
\bottomrule
\end{tabular}
\vspace{-0.5em}
\caption{Performance of GeoFMs across different training scenarios. Values in parentheses show scores when excluding a 3-pixel buffer around buildings, also during evaluation.}
\label{tab:geofms}
\end{table}

\begin{figure*}[htpb]
    \centering
    \includegraphics[width=\linewidth]{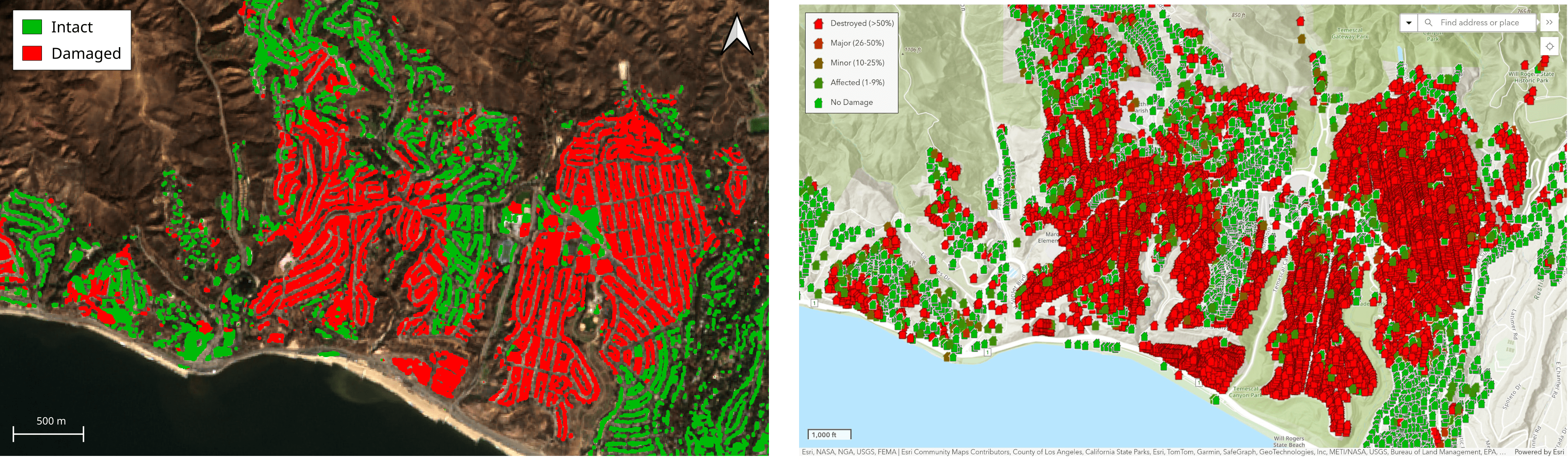}
    \vspace{-0.5em}
    \caption{\underline{Left:} output of our models when trained on the entire dataset. \underline{Right:} reference map adapted from the damage assessment conducted by the ~\cite{palisadeFireMaps}. Base maps: Sentinel-2 TCI / ESRI World Terrain Base.}
    \label{fig:california_fires}
\end{figure*}

\subsection{Inference on unseen disasters}
To test generalization to real-world examples outside of xBD-S12, we apply our trained U-Net model to the 2025 Palisades wildfire, a large recent event that occurred in January 2025 in the western suburbs of Los Angeles. We acquire Sentinel imagery following the same protocol as for xBD-S12, with post-event acquisitions on February 1 for Sentinel-2 and on February 7 for Sentinel-1.

\cref{fig:california_fires} compares our damage predictions with the official damage assessment from the ~\cite{palisadeFireMaps}. 
Although we cannot carry out a quantitative assessment, as the raw data is not publicly available, we observe an excellent correlation between our damage map and the dashboard visualizations of the official assessment.

\section{Conclusion}
\label{sec:conclusion}
We have shown that freely available medium-resolution satellite imagery from the Copernicus missions can support rapid damage assessment for disaster response. To that end, we have constructed xBD-S12, a dataset of paired pre- and post-event images from Sentinel-1 and Sentinel-2, with damage annotations derived from the earlier VHR xBD dataset.

We find that, in many scenarios, building damage is detectable despite the 10$\,$m GSD.
For disasters that induce large-scale, spectrally distinct changes, such as wildfires or floods, performance is consistently high. Events with more localized damage, such as hurricanes, are more challenging. For small, scattered damages like those from weak earthquakes, Copernicus data are not suitable due to insufficient resolution, at least under our current methodological framework.

Importantly, we do not regard moderate-resolution imagery as a replacement for VHR products, but rather as a complementary data source in the disaster response workflow. Especially in the absence of VHR imagery, damage assessment with Copernicus data can go a surprisingly long way. Its main value lies in rapid, coarse coverage, whereas a higher resolution is better suited for detailed damage inventories.

We have followed the task definition of xBD, but we note that, in operational settings, building footprints are often available from national inventories or global datasets such as Overture Maps~(\citeauthor{OvertureMaps}). Where such data exists, one would typically prefer it and employ satellite imagery only for damage classification. Especially at 10$\,$m GSD, building localization remains the weakest component of the pipeline.

Our approach deliberately prioritizes simplicity and accessibility, using only standard GRD products from Sentinel-1 and bi-temporal acquisitions. Future work could explore richer temporal information from longer time series~\citep[e.g.,][]{Dietrich2025}, or incorporate InSAR products such as coherence maps or damage proxy maps, which have shown great promise for detecting structural changes~\citep[e.g.,][]{ainscoe2025earthquake, Scher2025}. Although geospatial foundation models were of limited use in our experiments, we emphasize that they continue to improve, and future versions may well become effective tools for damage assessment.

{
    \begin{spacing}{1.02}
        \fontsize{8.5pt}{10pt}\selectfont
        \bibliography{references}

@article{van2006impacts,
  title={The impacts of climate change on the risk of natural disasters},
  author={Van Aalst, Maarten K},
  journal={Disasters},
  volume={30},
  number={1},
  pages={5--18},
  year={2006}
}

@incollection{banholzer2014impact,
  title={The impact of climate change on natural disasters},
  author={Banholzer, Sandra and Kossin, James and Donner, Simon},
  editor = {Z. Zommers and A. Singh},
  booktitle={Reducing disaster: Early warning systems for climate change},
  pages={21--49},
  year={2014},
  publisher={Springer}
}

@inbook{IPCC_2023,
    title={Weather and Climate Extreme Events in a Changing Climate},
    booktitle={Climate Change 2021 – The Physical Science Basis: Working Group I},
    publisher={Cambridge University Press},
    author={{IPCC}},
    year={2023},
    pages={1513–1766}
}

@article{kawasaki2013growing,
  title={The growing role of web-based geospatial technology in disaster response and support},
  author={Kawasaki, Akiyuki and Berman, Merrick Lex and Guan, Wendy},
  journal={Disasters},
  volume={37},
  number={2},
  pages={201--221},
  year={2013}
}

@article{Rolla2025,
  title = {Satellite‐Aided Disaster Response},
  volume = {6},
  number = {1},
  journal = {AGU Advances},
  author = {Rolla,  Julie and Khuller,  Aditya and An,  Karen and Emberson,  Robert and Fielding,  Eric and Schultz,  Lori and Miner,  Kimberley},
  year = {2025},
  month = feb 
}

@article{sun2020applications,
  title={Applications of artificial intelligence for disaster management},
  author={Sun, Wenjuan and Bocchini, Paolo and Davison, Brian D},
  journal={Natural Hazards},
  volume={103},
  number={3},
  pages={2631--2689},
  year={2020}
}

@article{braik2024automated,
  title={Automated building damage assessment and large-scale mapping by integrating satellite imagery, {GIS}, and deep learning},
  author={Braik, Abdullah M and Koliou, Maria},
  journal={Computer-Aided Civil Infrastruct.\ Eng.},
  volume={39},
  number={15},
  pages={2389--2404},
  year={2024}
}

@inproceedings{gupta2019creating,
  title={Creating {xBD}: A dataset for assessing building damage from satellite imagery},
  author={Gupta, Ritwik and Goodman, Bryce and Patel, Nirav and Hosfelt, Ricky and Sajeev, Sandra and Heim, Eric and Doshi, Jigar and Lucas, Keane and Choset, Howie and Gaston, Matthew},
  booktitle={Comput.\ Vis.\ Pat.\ Rec.\ Workshops},
  year={2019}
}

@article{gupta2019xbd,
    title={{xBD}: A dataset for assessing building damage from satellite imagery},
    author={Gupta, Ritwik and Hosfelt, Richard and Sajeev, Sandra and Patel, Nirav and Goodman, Bryce and Doshi, Jigar and Heim, Eric and Choset, Howie and Gaston, Matthew},
    journal={preprint arXiv:1911.09296},
    year={2019}
}

@inproceedings{fujita2017damage,
  title={Damage detection from aerial images via convolutional neural networks},
  author={Fujita, Aito and Sakurada, Ken and Imaizumi, Tomoyuki and Ito, Riho and Hikosaka, Shuhei and Nakamura, Ryosuke},
  booktitle={IAPR Conf.\ Machine Vision Appl.},
  year={2017}
}

@inproceedings{rudner2019multi3net,
  title={{Multi3Net}: segmenting flooded buildings via fusion of multiresolution, multisensor, and multitemporal satellite imagery},
  author={Rudner, Tim GJ and Ru{\ss}wurm, Marc and Fil, Jakub and Pelich, Ramona and Bischke, Benjamin and Kopa{\v{c}}kov{\'a}, Veronika and Bili{\'n}ski, Piotr},
  booktitle={AAAI Conf.\ on Artif.\ Intell.},
  year={2019}
}

@inproceedings{gu2024mamba,
  title={Mamba: Linear-time sequence modeling with selective state spaces},
  author={Gu, Albert and Dao, Tri},
  booktitle={Conf.\ on Language Modeling},
  year={2024}
}

@article{xu2019building,
  title={Building damage detection in satellite imagery using convolutional neural networks},
  author={Xu, Joseph Z and Lu, Wenhan and Li, Zebo and Khaitan, Pranav and Zaytseva, Valeriya},
  journal={preprint arXiv:1910.06444},
  year={2019}
}

@article{cao2020building,
  title={Building damage annotation on post-hurricane satellite imagery based on convolutional neural networks},
  author={Cao, Quoc Dung and Choe, Youngjun},
  journal={Natural Hazards},
  volume={103},
  number={3},
  pages={3357--3376},
  year={2020}
}

@article{galanis2021damagemap,
  title={{DamageMap}: A post-wildfire damaged buildings classifier},
  author={Galanis, Marios and Rao, Krishna and Yao, Xinle and Tsai, Yi-Lin and Ventura, Jonathan and Fricker, G Andrew},
  journal={Int.\ J.\ Disaster Risk Red.},
  volume={65},
  pages={102540},
  year={2021}
}

@misc{Durnov2019xview2,
  author = {Durnov, Victor},
  title = {{xView2}: 1st place solution},
  howpublished = {GitHub repository, \url{https://github.com/vdurnov/xview2_1st_place_solution}},
  note = {(accessed 2025-11-04)},
  year = {2019}
}

@inproceedings{zhao2020building,
  title={Building damage evaluation from satellite imagery using deep learning},
  author={Zhao, Fei and Zhang, Chengcui},
  booktitle={Int.\ Conf.\ Inform.\ Reuse \& Integ.\ Data Sci.},
  year={2020}
}

@article{wu2021building,
  title={Building damage detection using U-Net with attention mechanism from pre-and post-disaster remote sensing datasets},
  author={Wu, Chuyi and Zhang, Feng and Xia, Junshi and Xu, Yichen and Li, Guoqing and Xie, Jibo and Du, Zhenhong and Liu, Renyi},
  journal={Remote Sensing},
  volume={13},
  number={5},
  pages={905},
  year={2021}
}

@article{weber2020building,
  title={Building disaster damage assessment in satellite imagery with multi-temporal fusion},
  author={Weber, Ethan and Kan{\'e}, Hassan},
  journal={preprint arXiv:2004.05525},
  year={2020}
}

@inproceedings{gupta2021rescuenet,
  title={{RescueNet}: Joint building segmentation and damage assessment from satellite imagery},
  author={Gupta, Rohit and Shah, Mubarak},
  booktitle={Int\ Conf.\ Pat.\ Rec.},
  year={2021}
}

@inproceedings{hao2021attention,
  title={An attention-based system for damage assessment using satellite imagery},
  author={Hao, Hanxiang and Baireddy, Sriram and Bartusiak, Emily R and Konz, Latisha and LaTourette, Kevin and Gribbons, Michael and Chan, Moses and Delp, Edward J and Comer, Mary L},
  booktitle={Int.\ Geosci.\ Remote Sens.\ Symp.},
  year={2021}
}

@article{zheng2021building,
  title={Building damage assessment for rapid disaster response with a deep object-based semantic change detection framework: From natural disasters to man-made disasters},
  author={Zheng, Zhuo and Zhong, Yanfei and Wang, Junjue and Ma, Ailong and Zhang, Liangpei},
  journal={Remote Sens.\ Env.},
  volume={265},
  pages={112636},
  year={2021}
}

@article{liu2022novel,
  title={A novel attention-based deep learning method for post-disaster building damage classification},
  author={Liu, Chang and Sepasgozar, Samad ME and Zhang, Qi and Ge, Linlin},
  journal={Expert Syst.\ Appl.},
  volume={202},
  pages={117268},
  year={2022}
}

@inproceedings{chen2022dual,
  title={Dual-tasks siamese transformer framework for building damage assessment},
  author={Chen, Hongruixuan and Nemni, Edoardo and Vallecorsa, Sofia and Li, Xi and Wu, Chen and Bromley, Lars},
  booktitle={Int.\ Geosci.\ Remote Sens.\ Symp.},
  year={2022}
}

@article{kaur2023large,
  title={Large-scale building damage assessment using a novel hierarchical transformer architecture on satellite images},
  author={Kaur, Navjot and Lee, Cheng-Chun and Mostafavi, Ali and Mahdavi-Amiri, Ali},
  journal={Computer-Aided Civil Infrastruct.\ Eng.},
  volume={38},
  number={15},
  pages={2072--2091},
  year={2023}
}

@article{chen2024changemamba,
  title={{ChangeMamba}: Remote sensing change detection with spatio-temporal state space model},
  author={Chen, Hongruixuan and Song, Jian and Han, Chengxi and Xia, Junshi and Yokoya, Naoto},
  journal={IEEE T.\ Geosci.\ Remote Sens.},
  volume={62},
  pages={1--20},
  year={2024}
}

@article{chen2025bright,
  title={\textsc{Bright}: A globally distributed multimodal building damage assessment dataset with very-high-resolution for all-weather disaster response},
  author={Chen, Hongruixuan and Song, Jian and Dietrich, Olivier and Broni-Bediako, Clifford and Xuan, Weihao and Wang, Junjue and Shao, Xinlei and Wei, Yimin and Xia, Junshi and Lan, Cuiling and others},
  journal={Earth Syst.\ Sci.\ Data},
  year={2025}
}

@article{wang2025disasterm3,
  title={{DisasterM3}: A Remote Sensing Vision-Language Dataset for Disaster Damage Assessment and Response},
  author={Wang, Junjue and Xuan, Weihao and Qi, Heli and Liu, Zhihao and Liu, Kunyi and Wu, Yuhan and Chen, Hongruixuan and Song, Jian and Xia, Junshi and Zheng, Zhuo and others},
  journal={preprint arXiv:2505.21089},
  year={2025}
}

@article{szwarcman2024prithvi,
  title={{Prithvi-EO-2.0}: A versatile multi-temporal foundation model for {Earth} observation applications},
  author={Szwarcman, Daniela and Roy, Sujit and Fraccaro, Paolo and G{\'\i}slason, {\TH}orsteinn El{\'\i} and Blumenstiel, Benedikt and Ghosal, Rinki and de Oliveira, Pedro Henrique and Almeida, Joao Lucas de Sousa and Sedona, Rocco and Kang, Yanghui and others},
  journal={preprint arXiv:2412.02732},
  year={2024}
}

@misc{Clay2023,
  author = {{Clay Foundation}},
  title = {The {Clay} Foundation Model -- An open source {AI} model and interface for {Earth}},
  howpublished = {GitHub repository, \url{https://github.com/Clay-foundation/model}},
  note = {(accessed 2025-11-04)},
  year = {2023}
}

@misc{melissa2025,
title = {Jamaica: International support ‘crucial’ to hurricane recovery says {Guterres}},
author = {{United Nations}},
howpublished = {UN News, \url{https://news.un.org/en/story/2025/11/1166248}},
note = {(accessed 2025-11-04)},
year = 2025}

@misc{mckoy2025,
author = {Jillian McKoy},
title = {Death Count for 2025 {LA} County Wildfires Likely Higher than Records Show, {BU} Research Finds},
howpublished = {Boston University, \url{https://www.bu.edu/articles/2025/death-count-california-wildfires-higher-than-recorded/}},
note = {(accessed 2025-11-04)},
year = 2025
}

@misc{mishra2025,
  author = {{United Nations}},
  title = {‘{The} needs are huge’: Pakistan reels from floods as millions left homeless},
  howpublished = {UN News, \url{https://news.un.org/en/story/2025/09/1165864}},
  year = 2025
}

@article{xiong2024neural,
  title={Neural plasticity-inspired multimodal foundation model for earth observation},
  author={Xiong, Zhitong and Wang, Yi and Zhang, Fahong and Stewart, Adam J and Hanna, Jo{\"e}lle and Borth, Damian and Papoutsis, Ioannis and Saux, Bertrand Le and Camps-Valls, Gustau and Zhu, Xiao Xiang},
  journal={preprint arXiv:2403.15356},
  year={2024}
}

@inproceedings{astruc2025anysat,
  title={{AnySat}: One Earth Observation Model for Many Resolutions, Scales, and Modalities},
  author={Astruc, Guillaume and Gonthier, Nicolas and Mallet, Clement and Landrieu, Loic},
  booktitle={Comput.\ Vis.\ Pat.\ Rec.},
  year={2025}
}

@article{corley2024change,
  title={A change detection reality check},
  author={Corley, Isaac and Robinson, Caleb and Ortiz, Anthony},
  journal={preprint arXiv:2402.06994},
  year={2024}
}

@article{irvin2024teochat,
  title={{TEOChat}: A large vision-language assistant for temporal earth observation data},
  author={Irvin, Jeremy Andrew and Liu, Emily Ruoyu and Chen, Joyce Chuyi and Dormoy, Ines and Kim, Jinyoung and Khanna, Samar and Zheng, Zhuo and Ermon, Stefano},
  journal={preprint arXiv:2410.06234},
  year={2024}
}

@article{ainscoe2025earthquake,
  title={Earthquake damage mapped more comprehensively and accurately by radar satellites than optical imagery},
  author={Ainscoe, Eleanor A and Swaminathan, Rohini and Way, Lin and Modugno, Sirio and Chin, Shi Tong and Panta, Niroj and Crevoisier, Thierry and Yun, Sang-Ho},
  journal={Nat.\ Comm.\ Earth Env.},
  volume={6},
  number={1},
  pages={631},
  year={2025}
}

@article{DisasterAdaptiveNet25,
    title={{DisasterAdaptiveNet}: A robust network for multi-hazard building damage detection from very-high-resolution satellite imagery},
    author={Hafner, Sebastian and Gerard, Sebastian and Sullivan, Josephine and Ban, Yifang},
    journal={Int\. J.\ Appl.\ Earth Obs.\ Geoinf.},
    volume={143},
    year={2025},
    pages={104756}
}

@inproceedings{perez2018film,
  title={{FiLM}: Visual reasoning with a general conditioning layer},
  author={Perez, Ethan and Strub, Florian and De Vries, Harm and Dumoulin, Vincent and Courville, Aaron},
  booktitle={AAAI Conf.\ on Artif.\ Intell.},
  year={2018}
}

@inproceedings{CloudScorePlus,
    title={Comprehensive quality assessment of optical satellite imagery using weakly supervised video learning},
    author={Pasquarella, Valerie J and Brown, Christopher F and Czerwinski, Wanda and Rucklidge, William J},
    booktitle={Comput.\ Vis.\ Pat.\ Rec.},
    year={2023}
}

@article{Gorelick2017,
    title = {{Google Earth Engine}: Planetary-scale geospatial analysis for everyone},
    volume = {202},
    journal = {Remote Sens.\ Env.},
    author = {Gorelick, Noel and Hancher, Matt and Dixon, Mike and Ilyushchenko, Simon and Thau, David and Moore, Rebecca},
    year = {2017},
    pages = {18–27}
}

@misc{delprete2025phidown,
    author = {Del Prete, Roberto},
    title = {phidown: A Python Tool for Streamlined Data Downloading from {CDSE}},
    year = {2025},
    howpublished = {Zenodo, \url{https://doi.org/10.5281/zenodo.15332053}},
    note = {(accessed 2025-11-04)}
}

@article{sirko2023high,
  title={High-resolution building and road detection from {Sentinel-2}},
  author={Sirko, Wojciech and Brempong, Emmanuel Asiedu and Marcos, Juliana TC and Annkah, Abigail and Korme, Abel and Hassen, Mohammed Alewi and Sapkota, Krishna and Shekel, Tomer and Diack, Abdoulaye and Nevo, Sella and others},
  journal={preprint arXiv:2310.11622},
  year={2023}
}

@article{loshchilov2017decoupled,
  title={Decoupled weight decay regularization},
  author={Loshchilov, Ilya and Hutter, Frank},
  journal={preprint arXiv:1711.05101},
  year={2017}
}

@misc{emdat,
  author = {{Centre for Research on the Epidemiology of Disasters}},
  title = {{EM-DAT}: The Emergency Events Database},
  year = {2024},
  howpublished = {Universit{\'e} Catholique de Louvain, \url{https://www.emdat.be}},
  note = {(accessed 2025-11-04)}
}

@article{gomes2025terratorch,
  title={{TerraTorch}: The Geospatial Foundation Models Toolkit},
  author={Gomes, Carlos and Blumenstiel, Benedikt and de Sousa Almeida, Joao Lucas and de Oliveira, Pedro Henrique and Fraccaro, Paolo and Escofet, Francesc Marti and Szwarcman, Daniela and Simumba, Naomi and Kienzler, Romeo and Zadrozny, Bianca},
  journal={preprint arXiv:2503.20563},
  year={2025}
}

@article{Stewart_TorchGeo_Deep_Learning_2024,
    author = {Stewart, Adam J. and Robinson, Caleb and Corley, Isaac A. and Ortiz, Anthony and Lavista Ferres, Juan M. and Banerjee, Arindam},
    doi = {10.1145/3707459},
    journal = {ACM Trans.\ Spatial Alg. \& Systems},
    month = dec,
    title = {{TorchGeo}: Deep Learning With Geospatial Data},
    year = {2024}
}

@inproceedings{xiao2018unified,
  title={Unified perceptual parsing for scene understanding},
  author={Xiao, Tete and Liu, Yingcheng and Zhou, Bolei and Jiang, Yuning and Sun, Jian},
  booktitle={Europ.\ Conf.\ Comput.\ Vis},
  year={2018}
}

@misc{palisadeFireMaps,
  author = {{California Dept.\ of Forestry and Fire Protection}},
  title = {Palisades Fire Damage Maps},
  year = {2025},
  howpublished = {\url{https://recovery.lacounty.gov/palisades-fire/}},
  note = {(accessed 2025-11-04)}
}

@misc{OvertureMaps,
    author = {{Overture Maps Foundation}},
    title = {{Open Map Data}},
    howpublished = {\url{https://overturemaps.org}},
    note = {(accessed 2025-11-04)}
}

@article{Dietrich2025,
  title = {An open-source tool for mapping war destruction at scale in Ukraine using Sentinel-1 time series},
  volume = {6},
  number = {1},
  journal = {Nat.\ Comm.\ Earth Env.},
  author = {Dietrich, Olivier and Peters, Torben and Sainte Fare Garnot,  Vivien and Sticher, Valerie and Ton-That Whelan, Thao and Schindler,  Konrad and Wegner, Jan Dirk},
  year = {2025}
}

@article{Scher2025,
  title = {Nationwide conflict damage mapping with interferometric synthetic aperture radar: A study of the 2022 {Russia}--{Ukraine} conflict},
  volume = {11},
  journal = {Sci. Remote Sensing},
  author = {Scher, Corey and Van Den Hoek, Jamon},
  year = {2025},
  pages = {100217}
}
	\end{spacing}
}

\onecolumn
\clearpage
\begin{appendices}

\section{xBD georeference correction}
\label{app:align_xbd}

While the original xBD images were distributed as standard PNG files for the xView2 competition, a subsequent release provided georeferenced raster versions. In this release, building polygons are supplied with coordinates in both image space ($xy$ pixel coordinates) and geographical space (longitude–latitude). Upon inspection, we observed that the $xy$ coordinates align perfectly with the images, as in the original challenge dataset, but the \emph{lon–lat} coordinates exhibit systematic misalignment. Specifically, the building footprints are correctly georeferenced, whereas the raster images appear shifted, likely due to a projection error during the georeferencing of the original PNG files. Since we need the correct image extents to download the corresponding Sentinel images, it was essential to correct for this misalignment.

To do this, we leveraged the fact that the building footprint coordinates are correct in both coordinate systems and used them as ground control points (GCPs) to register the images. This approach, however, can only be applied to patches containing enough buildings, and many tiles do not contain any. Our analysis revealed that all tiles derived from the same original MAXAR image share a consistent spatial shift, suggesting that a global affine transformation per MAXAR image could achieve the correction for all tiles.

Our correction procedure consists of three steps: (i) we compute an affine transformation $\mathbf{A}_i$ for each tile $i$ containing sufficient GCPs; (ii) we derive a robust global transformation $\mathbf{A}_{\text{global}}$ by taking the element-wise median across all individual transforms, and (iii) we correct residual offsets by enforcing that GCPs located on image edges (i.e., with $x$- or $y$-coordinates equal to 0 or 1024) map exactly to the tile boundaries. This global transformation is then applied to all tiles to correct the spatial misalignment. 

\begin{figure}[!htpb]
    \centering
    \includegraphics[width=.9\textwidth]{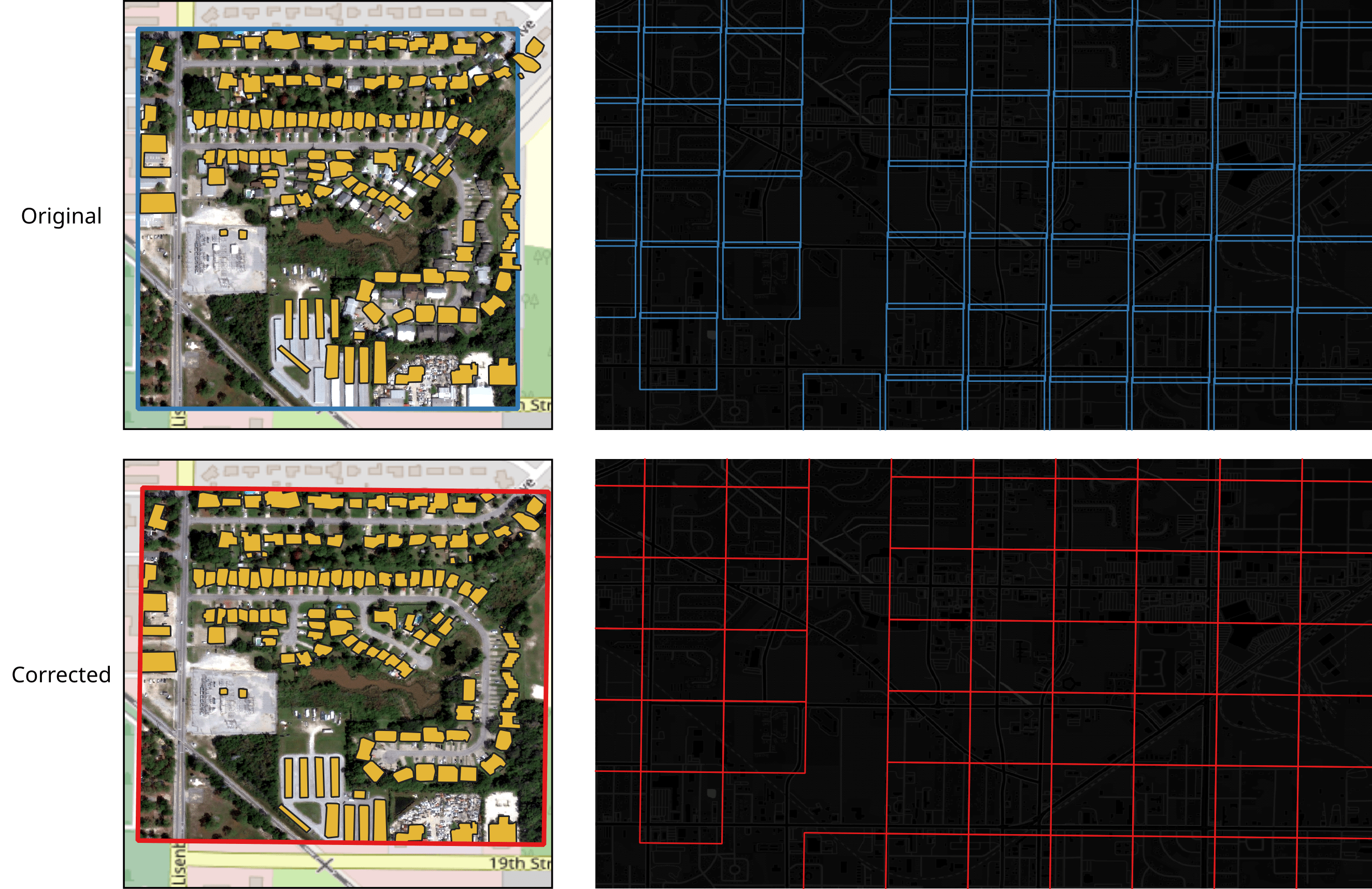}
    \caption{Comparison of original (\squarecolor{3579b1}) and corrected (\squarecolor{e41a1c}) image georeferencing. \underline{Left:} Georeferenced overlay of building footprints on VHR satellite imagery for a single patch, showing original alignment (top) and corrected alignment (bottom). \underline{Right:} Spatial distribution of image extents before (top) and after (bottom) correction. Both examples are from hurricane Michael. The corrected version successfully recovers the original regular grid structure. Basemaps: OpenStreetMap/CartoDB Dark Matter.}
    \label{fig:alignment}
\end{figure}

While the described approach is somewhat heuristic, it successfully recovers the original regular grid used to partition MAXAR images into 1024×1024 patches, as demonstrated in \cref{fig:alignment}. The corrected georeferencing ensures accurate spatial alignment between the xBD dataset and auxiliary satellite imagery sources.

\clearpage
\section{Details on Sentinel tile selection}
\label{app:sentinel_details}

\begin{figure}[!htpb]
    \centering
    \includegraphics[width=\textwidth]{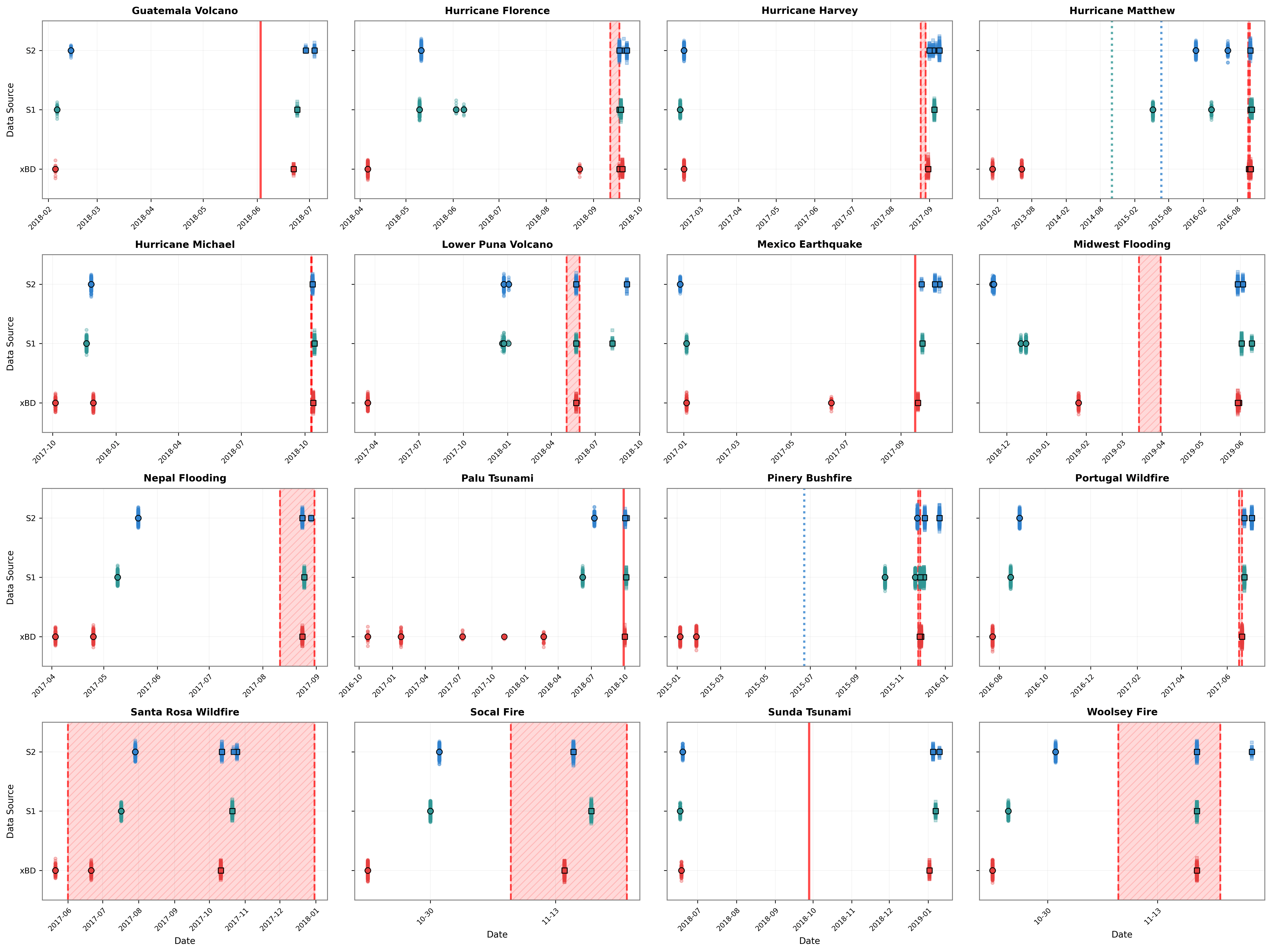}
    \caption{Temporal distribution of imagery acquisitions across disasters. Each subplot shows the acquisition dates for xBD (\squarecolor{E53E3E}), Sentinel-1 (\squarecolor{319795}), and Sentinel-2 (\squarecolor{3182CE}) imagery, with circles denoting pre-disaster and squares post-disaster acquisitions. Vertical jitter is applied to visualize the number of tiles per date. Red shaded regions indicate disaster periods (from EM-DAT~\citep{emdat} via \cite{DisasterAdaptiveNet25}); single red lines mark disasters with one temporal reference. Dotted vertical lines show Sentinel-1 (turquoise) and Sentinel-2 (blue) launch dates where relevant.}
    \label{fig:sentinel_dates_distribution}
\end{figure}

We selected Sentinel-2 and Sentinel-1 imagery at the patch level, meaning that patches for the same disaster may originate from different Sentinel tiles, e.g. due to varying cloud coverage. \cref{fig:sentinel_dates_distribution} summarizes all tiles used per disaster event.

\paragraph{Sentinel-2 selection.}
We manually preselected candidate dates based on global cloud coverage and post-event damage visibility. When multiple candidates were available, we computed the average cloud score $CS$ per patch using Google's CloudScore+~\citep{CloudScorePlus}. For pre-event imagery, we selected the date maximizing $CS$. For post-event imagery, we balanced cloud coverage and temporal proximity using a custom \textit{cloud-temporal score}:
\begin{equation}
    \text{date}_{\text{post}} = \text{argmax}_i \left(\alpha_{\text{cloud}} \cdot CS_i + \alpha_{\text{time}} \cdot TS_i\right), \quad \text{where} \quad TS_i = \frac{1}{\sqrt{i + 1}},
\end{equation}
with $\alpha_{\text{cloud}}=0.4$, $\alpha_{\text{time}}=0.6$ (determined empirically), and $i$ the temporal index of the candidates (0 for the first post-disaster candidate, 1 for the second, etc.). We added a hard threshold to exclude candidates with $CS_i \leq 0.5$ (respectively $CS_i \leq 0.35$ for the Lower Puna Volcano). If none remained, we selected the highest-scoring date without this constraint. Despite our efforts, some post-event imagery might still be cloudy.

\paragraph{Sentinel-1 selection.}
We then selected the Sentinel-1 images temporally closest to the chosen Sentinel-2 dates, while ensuring that both pre- and post-event imagery share the same orbital paths and that the tiles are fully valid (after mosaicking). For Hurricane Matthew, we were unable to find an orbital path that met our criteria, so we sampled images from two different orbits.

\clearpage
\section{Ablation of data input}
\label{app:ablation_data_input}
In this section, we examine the separate contributions of Sentinel-1 and Sentinel-2 to the overall performance. Results are reported in \cref{tab:score_data_input}.

Across both splits, Sentinel-2 alone significantly outperforms Sentinel-1 alone, suggesting that optical imagery provides richer spectral information about surface changes than SAR amplitude data. However, combining both modalities consistently improves over either input alone, which confirms our initial intuition that Sentinel-1 and Sentinel-2 provide complementary information.

Interestingly, the added value of Sentinel-1 is more pronounced in the event-based split, where it raises the overall F1\textsubscript{comp} from 0.677 to 0.710, compared to a marginal gain from 0.756 to 0.761 in the xView2 split. This asymmetry suggests that, in the xView2 split setting, the model relies heavily on event-specific spectral signatures visible in the Sentinel-2 optical imagery, which leaves little room for SAR to contribute further. In contrast, when generalizing to unseen disasters, the model relies more on complementary information from both data sources, and adding Sentinel-1 data brings more benefits.

A notable exception is the Mexico earthquake, for which Sentinel-1 alone achieves the highest score. However, as discussed in \cref{app:poor_performing_disasters}, this disaster is an extreme outlier with near-zero damage detection performance across all configurations, so this result should not be over-interpreted.

\begin{table}[!htpb]
\centering
\footnotesize
\begin{tabular}{@{}l|lccc@{}}
\toprule
 \multicolumn{1}{c}{} 
& & only S1 & only S2 & S1 + S2 \\
\midrule
\multirow{11}{*}{\rotatebox{90}{\textit{xView2 split}}}
& Guatemala volcano & 0.348 {\scriptsize (0.351)} & 0.566 {\scriptsize (0.636)} & \textbf{0.611} {\scriptsize \textbf{(0.671)}} \\
& Hurricane Florence & 0.503 {\scriptsize (0.522)} & \textbf{0.785} {\scriptsize (0.848)} & \textbf{0.785} {\scriptsize \textbf{(0.852)}} \\
& Hurricane Harvey & 0.447 {\scriptsize (0.513)} & 0.743 {\scriptsize (0.832)} & \textbf{0.749} {\scriptsize \textbf{(0.839)}} \\
& Hurricane Matthew & 0.288 {\scriptsize (0.327)} & \textbf{0.557} {\scriptsize (0.625)} & \textbf{0.557} {\scriptsize \textbf{(0.628)}} \\
& Hurricane Michael & 0.435 {\scriptsize (0.469)} & 0.586 {\scriptsize (0.672)} & \textbf{0.596} {\scriptsize \textbf{(0.683)}} \\
& Mexico earthquake & \textbf{0.259} {\scriptsize \textbf{(0.337)}} & 0.194 {\scriptsize (0.287)} & 0.194 {\scriptsize (0.287)} \\
& Midwest flooding & 0.388 {\scriptsize (0.414)} & 0.588 {\scriptsize (0.664)} & \textbf{0.612} {\scriptsize \textbf{(0.692)}} \\
& Palu tsunami & 0.442 {\scriptsize (0.510)} & 0.771 {\scriptsize (0.857)} & \textbf{0.777} {\scriptsize \textbf{(0.863)}} \\
& Santa Rosa wildfire & 0.544 {\scriptsize (0.602)} & \textbf{0.842} {\scriptsize (0.934)} & \textbf{0.842} {\scriptsize \textbf{(0.942)}} \\
& Socal fire & 0.408 {\scriptsize (0.453)} & 0.741 {\scriptsize (0.817)} & \textbf{0.763} {\scriptsize \textbf{(0.843)}} \\[0.5ex]
\cline{2-5}
& \rule{0pt}{3ex} Overall & 0.478 {\scriptsize (0.537)} & 0.756 {\scriptsize (0.842)} & \textbf{0.761} {\scriptsize \textbf{(0.849)}} \\
\midrule
\multirow{6}{*}{\rotatebox{90}{\textit{event-based}}}
& Guatemala volcano & 0.355 {\scriptsize (0.373)} & 0.650 {\scriptsize (0.733)} & \textbf{0.721} {\scriptsize \textbf{(0.809)}} \\
& Hurricane Matthew & 0.207 {\scriptsize (0.224)} & 0.363 {\scriptsize (0.431)} & \textbf{0.403} {\scriptsize \textbf{(0.474)}} \\
& Nepal flooding & 0.337 {\scriptsize (0.337)} & 0.563 {\scriptsize (0.582)} & \textbf{0.616} {\scriptsize \textbf{(0.655)}} \\
& Santa Rosa wildfire & 0.469 {\scriptsize (0.521)} & 0.700 {\scriptsize (0.777)} & \textbf{0.747} {\scriptsize \textbf{(0.828)}} \\
& Sunda tsunami & 0.200 {\scriptsize (0.250)} & 0.283 {\scriptsize (0.332)} & \textbf{0.336} {\scriptsize \textbf{(0.396)}} \\[0.5ex]
\cline{2-5}
& \rule{0pt}{3ex} Overall & 0.399 {\scriptsize (0.428)} & 0.677 {\scriptsize (0.734)} & \textbf{0.710} {\scriptsize \textbf{(0.775)}} \\
\bottomrule
\end{tabular}
\vspace{-0.5em}
\caption{Per-disaster and global F1\textsubscript{comp} score when trained on different data modalities and for both test splits. As in \cref{tab:main_results}, values in parentheses are computed excluding a 3-pixel buffer around buildings during evaluation.}
\label{tab:score_data_input}
\end{table}

\clearpage
\section{Analysis of outlier disasters}
\label{app:poor_performing_disasters}
In this section, we examine the two disasters for which performance differs the most from the general trend: the Guatemala volcano eruption and the Mexico earthquake.

\begin{figure}[!htpb]
    \centering
    \includegraphics[width=.9\textwidth]{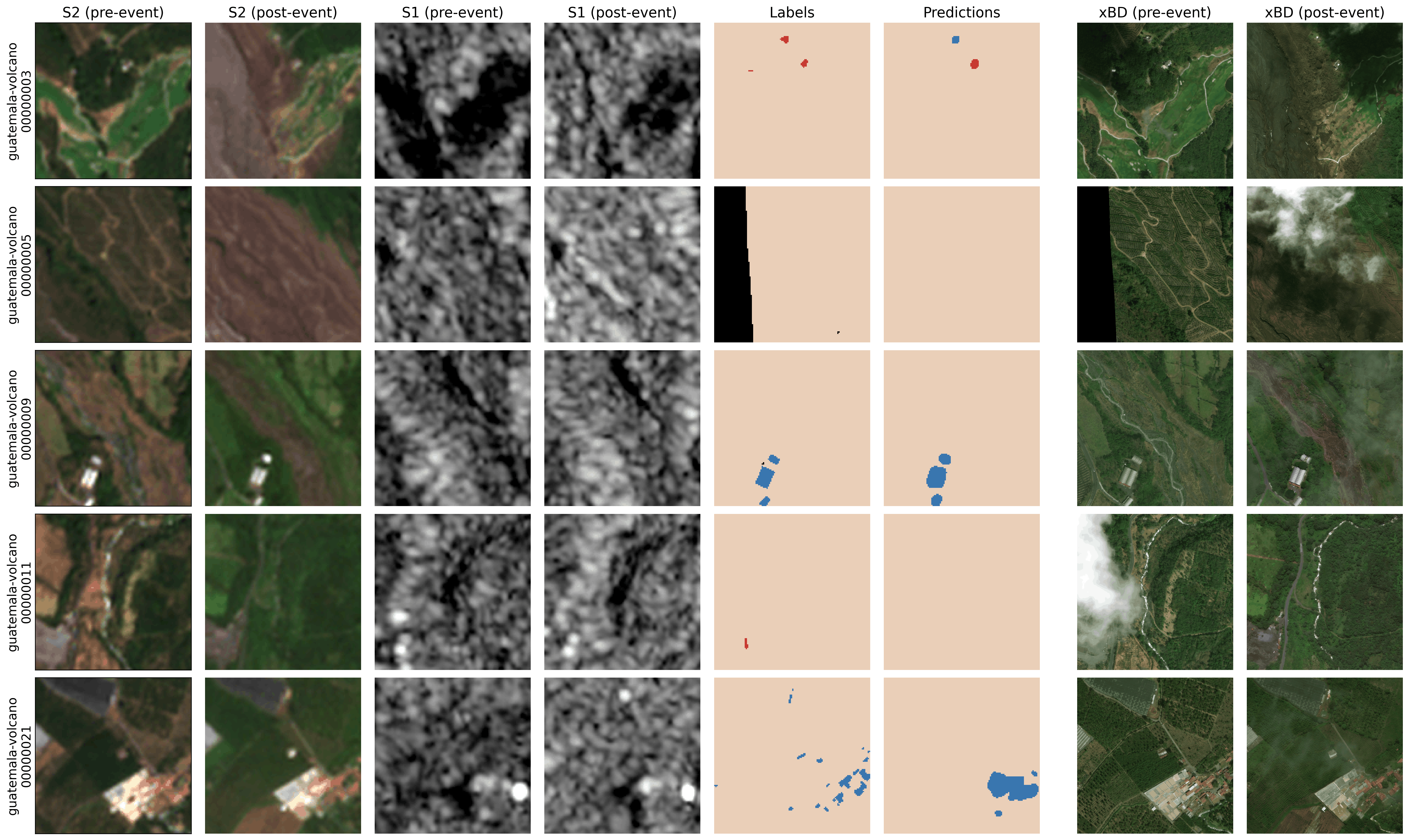}\\ \vspace{-0.5em}
    {\footnotesize
    \squarecolor{eacfb8} background\quad\squarecolor{000000} masked\quad\squarecolor{3976af} intact\quad\squarecolor{c73a31} damaged
    }
    \caption{Complete test set for the Guatemala volcano disaster in the original xView2 split and our predictions. xBD images are shown for reference.}
    \label{fig:guatemala_full_test}
\end{figure}

\paragraph{Guatemala volcano.} The xView2 test set for this disaster contains only 5 patches, making it particularly sensitive to the content of these patches. Upon visual inspection (\cref{fig:guatemala_full_test}), the few damaged buildings are either small and isolated or located at the boundary of lava flows, both of which are challenging at 10$\,$m GSD. Additionally, the last patch contains a large farm, which represents a significant distribution shift from the rest of the dataset, and further confuses the model. Together, these factors explain the counterintuitive finding that performance is higher on the event-based split for this disaster. 

\paragraph{Mexico earthquake.} Unlike the Guatemala case, this disaster exposes a genuine limitation of medium-resolution imagery rather than an artifact of test set composition. The affected area is a dense urban environment where damage is both spatially sparse and of low intensity, with only a handful of individual buildings damaged among otherwise intact city blocks. Such damage patterns are extremely difficult to detect at Sentinel's 10$\,$m GSD, and show the limits of our approach when compared to VHR images.

\begin{figure}[htpb]
    \centering
    \includegraphics[width=\textwidth]{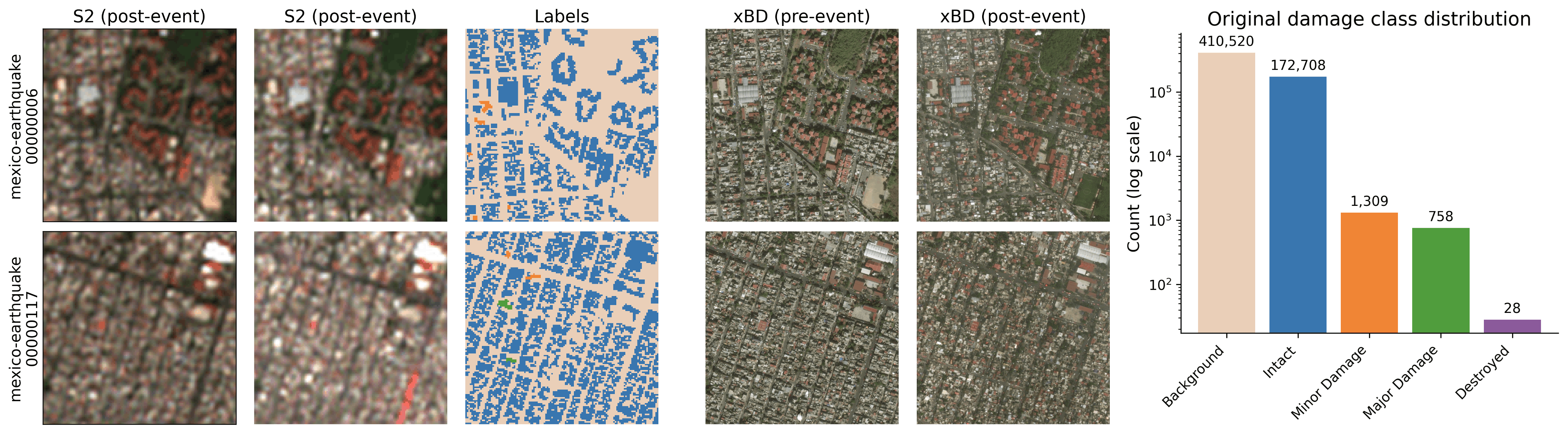}
    \caption{\textbf{Left:} Sample patches from the test set containing damaged buildings. xBD images are shown for reference. Sentinel-2 does not offer the resolution to visually identify isolated, low-intensity damage within largely intact urban structures. \textbf{Right:} Original damage class distribution in the test set (log scale).}
    \label{fig:mexico}
\end{figure}


\clearpage
\section{Additional visualization}
\label{app:additional_visu}
\begin{figure}[!htpb]
    \centering
    \includegraphics[width=\textwidth]{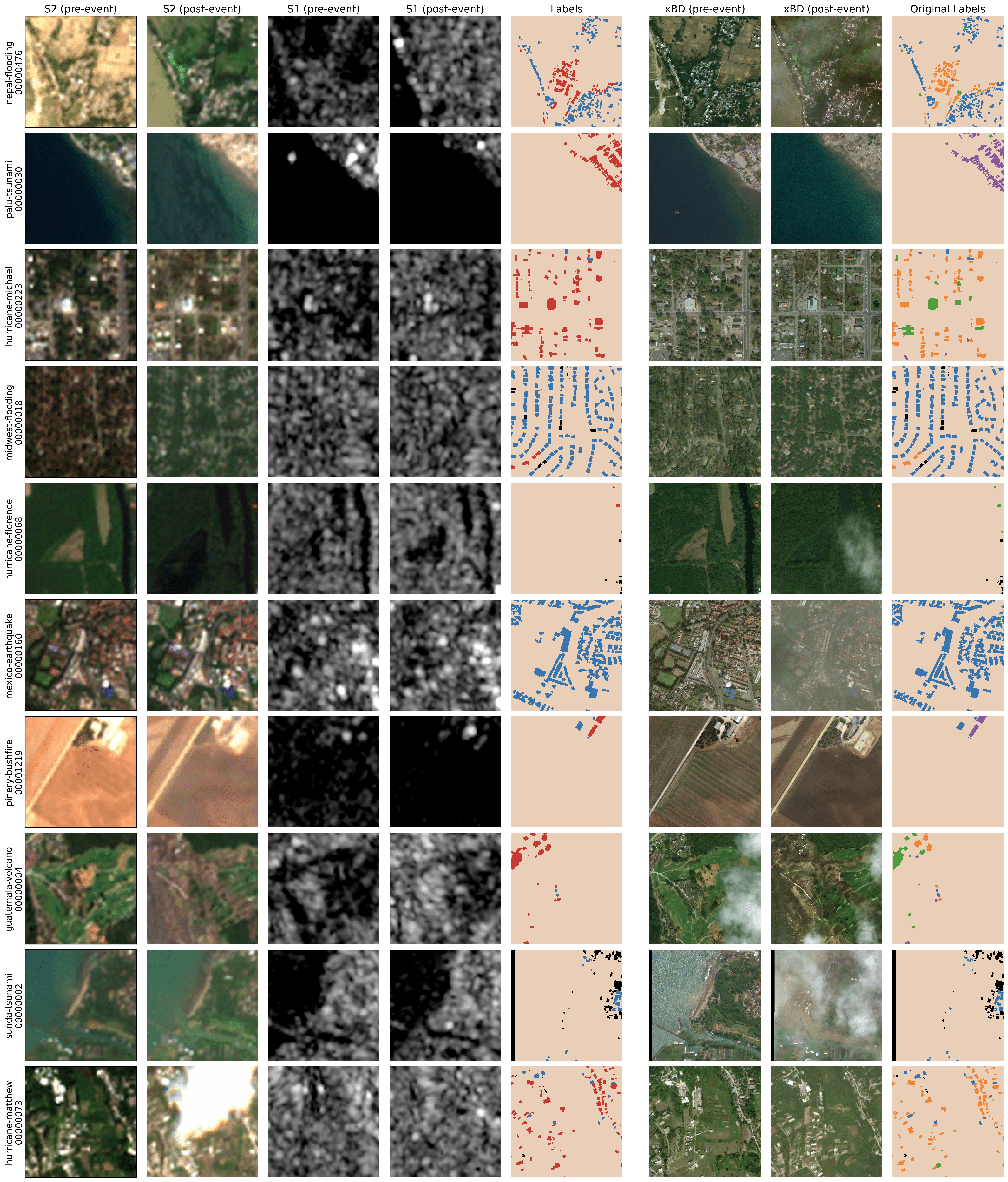}
\end{figure}
\begin{figure}[!htpb]
    \centering
    \includegraphics[width=\textwidth]{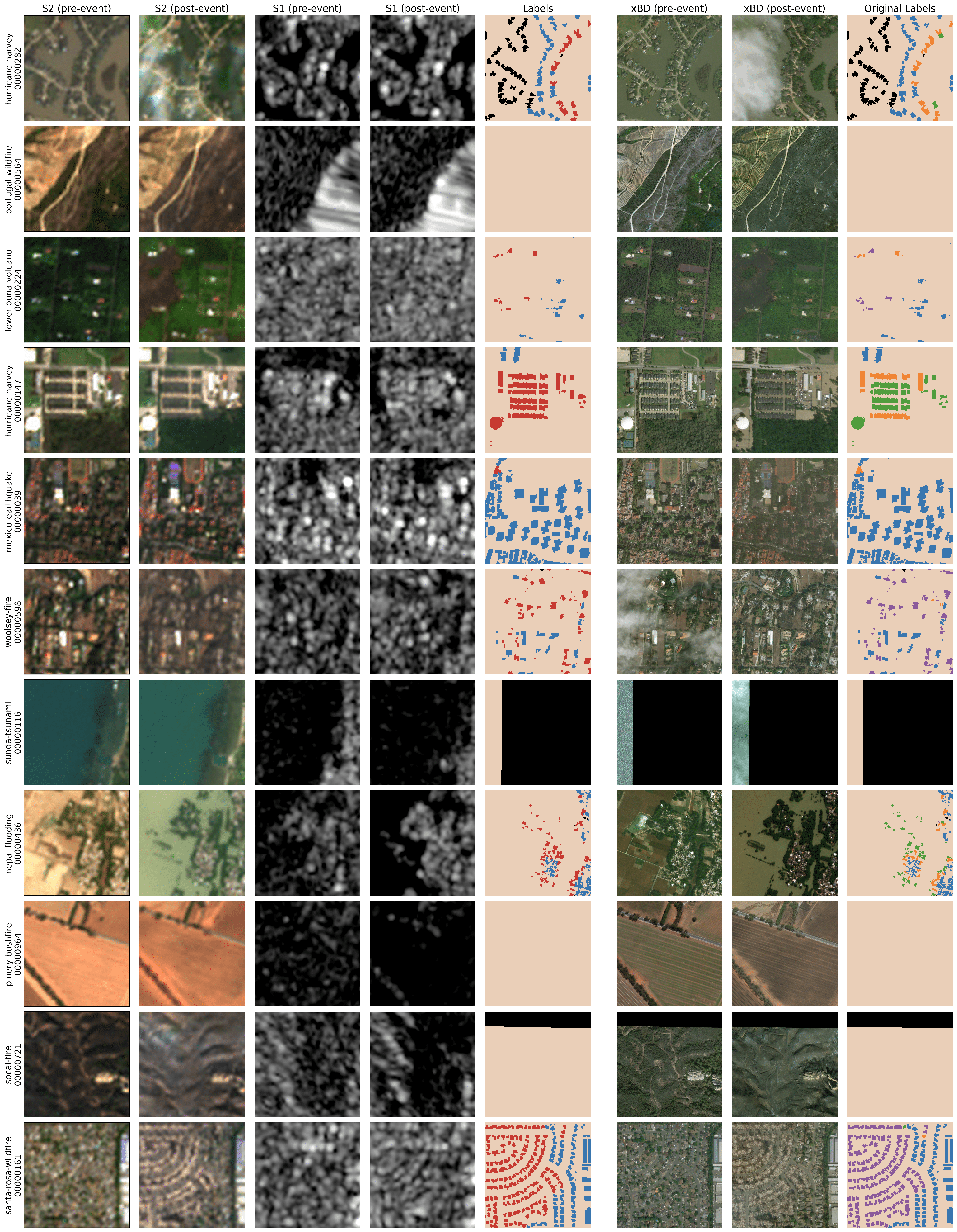}\\
    {\footnotesize
    \squarecolor{eacfb8} background\quad\squarecolor{000000} unknown\quad\squarecolor{3976af} undamaged\quad\squarecolor{c73a31} damaged\quad(\quad\squarecolor{f08535} \textcolor{gray}{minor damage}\quad\squarecolor{509d3d} \textcolor{gray}{major damage}\quad\squarecolor{8B5A9B} \textcolor{gray}{destroyed}\quad)
    }
    \caption{Additional example patches from xBD-S12. For visualisation purposes, we display the True Color Image product for Sentinel-2 and the VV-polarised (log-)amplitude for Sentinel-1. All tiles are 128$\times$128$\,$px ($\approx$4$\,$m GSD). On the right, VHR images (1024$\times$1024$\,$px, $\approx$0.5$\,$m GSD) and labels from the original high-resolution xBD dataset are shown for reference.}
    \label{fig:additional_visu}
\end{figure}

\clearpage
\section{Details on geospatial foundation model finetuning}
\label{app:geofm_finetuning}

To evaluate whether pretrained geospatial foundation models (GeoFMs) can improve damage assessment, we test two recent architectures: Prithvi-EO-2.0-300M~\citep{szwarcman2024prithvi} and DOFA-Base~\citep{xiong2024neural}. For both models, we freeze the encoder backbone and finetune only a UperNet~\citep{xiao2018unified} decoder head. Training follows the protocol described in \cref{sec:training_details}, except that we reduce the number of epochs to 20 and only conduct one run for each model instead of the usual ensemble of three runs.

\paragraph{Prithvi-EO-2.0-300M.}
We use the pretrained model from TerraTorch~\citep{gomes2025terratorch}. Following its pretraining protocol, we select six Sentinel-2 bands (B2, B3, B4, B8, B11, B12) and exclude Sentinel-1 data. Input images are standardized using the statistics provided by the model authors and resized to $224 \times 224$ pixels via bilinear interpolation. Since Prithvi natively processes temporal stacks, we directly concatenate the pre- and post-event images along the temporal dimension. The decoder receives multi-scale features extracted from encoder blocks \{5, 11, 17, 23\}.

\paragraph{DOFA-Base.}
We use the pretrained model from TorchGeo~\citep{Stewart_TorchGeo_Deep_Learning_2024}. DOFA's pretraining incorporates both Sentinel-1 and Sentinel-2 imagery. For Sentinel-1, we convert log-amplitude values back to linear amplitude $x' = 10^{x/10}$, and clip to [0, 1]. For Sentinel-2, we construct a 9-channel input that matches the pretraining configuration: the first three channels contain the RGB bands from the True Color Image product, followed by bands \{B5, B6, B7, B8, B11, B12\}, scaled by dividing by 32 and clipping to 255. All inputs are standardized using the statistics from the official implementation and resized to $224 \times 224$ pixels. Unlike Prithvi, DOFA does not process temporal sequences jointly. We therefore encode the pre- and post-event images independently through the frozen backbone, concatenate the resulting feature maps channel-wise, and feed them to the decoder. We extract features from encoder blocks \{3, 6, 9, 11\}.

\end{appendices}
\end{document}